\documentclass[12pt]{article}      %
\usepackage[margin=1in]{geometry}  %
\usepackage{lineno}

\usepackage[english]{babel}

\usepackage{amsmath,amssymb,amsthm}  %
\usepackage{mathtools}               %

\usepackage{graphicx}
\usepackage{subcaption}             %
\usepackage{float}                  %

\usepackage{booktabs}   %
\usepackage{siunitx}    %

\usepackage[hidelinks]{hyperref}      %
\usepackage[capitalise]{cleveref}     %
\usepackage[numbers]{natbib}          %
\usepackage[numbers]{natbib}   %
\usepackage{setspace}
\theoremstyle{plain}

\theoremstyle{definition}

\makeatletter            %
\newif\ifistwocolumn     %
\if@twocolumn            %
  \istwocolumntrue
\else
  \istwocolumnfalse
\fi
\makeatother

\usepackage{newtxtext,newtxmath}

\title{\vspace{-3mm}Improving Generative Ad Text on Facebook using Reinforcement Learning}

\author{\large Daniel R. Jiang$^{\dagger,*}$\!,\, Alex Nikulkov$^{\dagger}$\!,\, Yu-Chia Chen,\, Yang Bai,\, Zheqing Zhu\vspace{10pt}\\ 
Meta Platforms, Menlo Park, California, USA.\vspace{10pt}\\
\small $^\dagger$These authors contributed equally to this work.\\
\small $^\ast$Corresponding author. Email: drjiang@meta.com}
\date{}

\begin{document}
\maketitle

\begin{abstract}
Generative artificial intelligence (AI), in particular large language models (LLMs), is poised to drive transformative economic change. LLMs are \emph{pre-trained} on vast text data to learn general language patterns, but a subsequent \emph{post-training} phase is critical to align them for specific real-world tasks. Reinforcement learning (RL) is the leading post-training technique, yet its economic impact remains largely underexplored and unquantified. We examine this question through the lens of the first deployment of an RL-trained LLM for generative advertising on Facebook. Integrated into Meta's \emph{Text Generation} feature, our model, ``AdLlama,'' powers an AI tool that helps advertisers create new variations of human-written ad text. To train this model, we introduce \emph{reinforcement learning with performance feedback} (RLPF), a post-training method that uses historical ad performance data as a reward signal. In a large-scale 10-week A/B test on Facebook spanning nearly 35{,}000 advertisers and 640{,}000 ad variations, we find that AdLlama improves click-through rates by 6.7\% ($p = 0.0296$) compared to a supervised imitation model trained on curated ads. This represents a substantial improvement in advertiser return on investment on Facebook. We also find that advertisers who used AdLlama generated more ad variations, indicating higher satisfaction with the model's outputs. To our knowledge, this is the largest study to date on the use of generative AI in an ecologically valid setting, offering an important data point quantifying the tangible impact of RL post-training. Furthermore, the results show that RLPF is a promising and generalizable approach for \emph{metric-driven post-training} that bridges the gap between highly capable language models and tangible outcomes.

\end{abstract}

\ifistwocolumn
\section*{Introduction}
\else
\section{Introduction}
\fi

\noindent Generative artificial intelligence (AI) is increasingly being recognized for its transformative potential across industries, driving innovations in content creation \cite{ramesh2021zero,chatgpt2022,lee2022coauthor,achiam2023gpt,zhou2024generative}, education \cite{kumar2023math, bastani2024generative,kreijkes2025effects}, medicine \cite{thirunavukarasu2023large,clusmann2023future,van2024adapted,sellergren2025medgemma},
and complex decision-making \cite{meta2022human,bubeck2023sparks,alphaproof2024}. 
It is also widely believed to have the potential for significant economic impact \cite{brynjolfsson2022turing,butler2023microsoft,hoffmann2024generative,handa2025economic,ju2025collaborating}, with pioneering work showing quantifiable benefits in the areas of customer support \cite{brynjolfsson2023generative,ni2024generative}, enterprise information tasks \cite{cambon2023early}, legal tasks \cite{choi2023ai,schwarcz2025ai}, consulting work \cite{dell2023navigating}, software development \cite{peng2023impact,cui2024effects,hoffmann2024generative}, email marketing \cite{angelopouloscausal}, online advertising \cite{chen2024large}, and a variety of professional writing tasks \cite{noy2023experimental}. However, turning this potential into consistent real-world impact depends critically on how models are fine-tuned and aligned during a phase of model training called \emph{post-training}.

The initial phase for large language models (LLM) training is called \emph{pre-training}, where the model learns general language patterns and world knowledge from large-scale unlabeled text data. However, to perform effectively in real-world applications, generative AI systems require some form of \emph{post-training}---a suite of methods designed to adapt pre-trained foundation models to specific downstream tasks \cite{lambert2024t}.
A common approach involves supervised fine-tuning (SFT), in which models are trained to follow human instructions by mimicking curated target responses \cite{ouyang2022training, zhou2023instruction}. Another example is the use of \emph{reinforcement learning from human feedback} (RLHF) to train the model on human preference data \cite{ziegler2019fine}, which has been shown to be highly effective in domains like summarization \cite{stiennon2020learning}, chatbot assistants \cite{bai2022training}, and instruction-following \cite{dubois2024alpacafarm}. More recently, researchers have found that reinforcement learning (RL) using \emph{verifiable rewards} (i.e., objective signals from domains like math or coding that can be automatically validated) can lead to marked improvements in LLM reasoning and problem-solving capabilities \cite{lambert2024t,guo2025deepseek}. 

\looseness-1 As noted above, there is a broadening body of research focused on quantifying the  impact of LLMs across various sectors. However, the specific impact of \emph{the post-training phase} has been relatively underexplored, despite its critical role in the development of virtually all prominent LLM models. In this paper, we present new findings on the concrete impact of RL for post-training. Our analysis is conducted through the perspective of the online advertising industry, which is a major driver of the global economy: worldwide expenditure in online ads is projected to reach \$513 billion USD in 2025 \cite{Dentsu2024} and is expected to represent 63\% of the total global advertising revenue across all mediums \cite{Dentsu2024b}.

Specifically, we focus on Meta's Text Generation product, which uses an LLM to generate variations of an advertiser's human-written ad text. This allows each advertiser to select multiple versions of the ad to show to potential customers, taking advantage of Meta's ad delivery systems to show the most performant variants. The initial version of the Text Generation product used an LLM that was trained in a naive manner, by fine-tuning it to imitate the style of a set of curated ads using supervised learning (SFT).

\ifistwocolumn
\begin{figure*}[!ht] %
	\centering
	\includegraphics[width=0.9\textwidth]{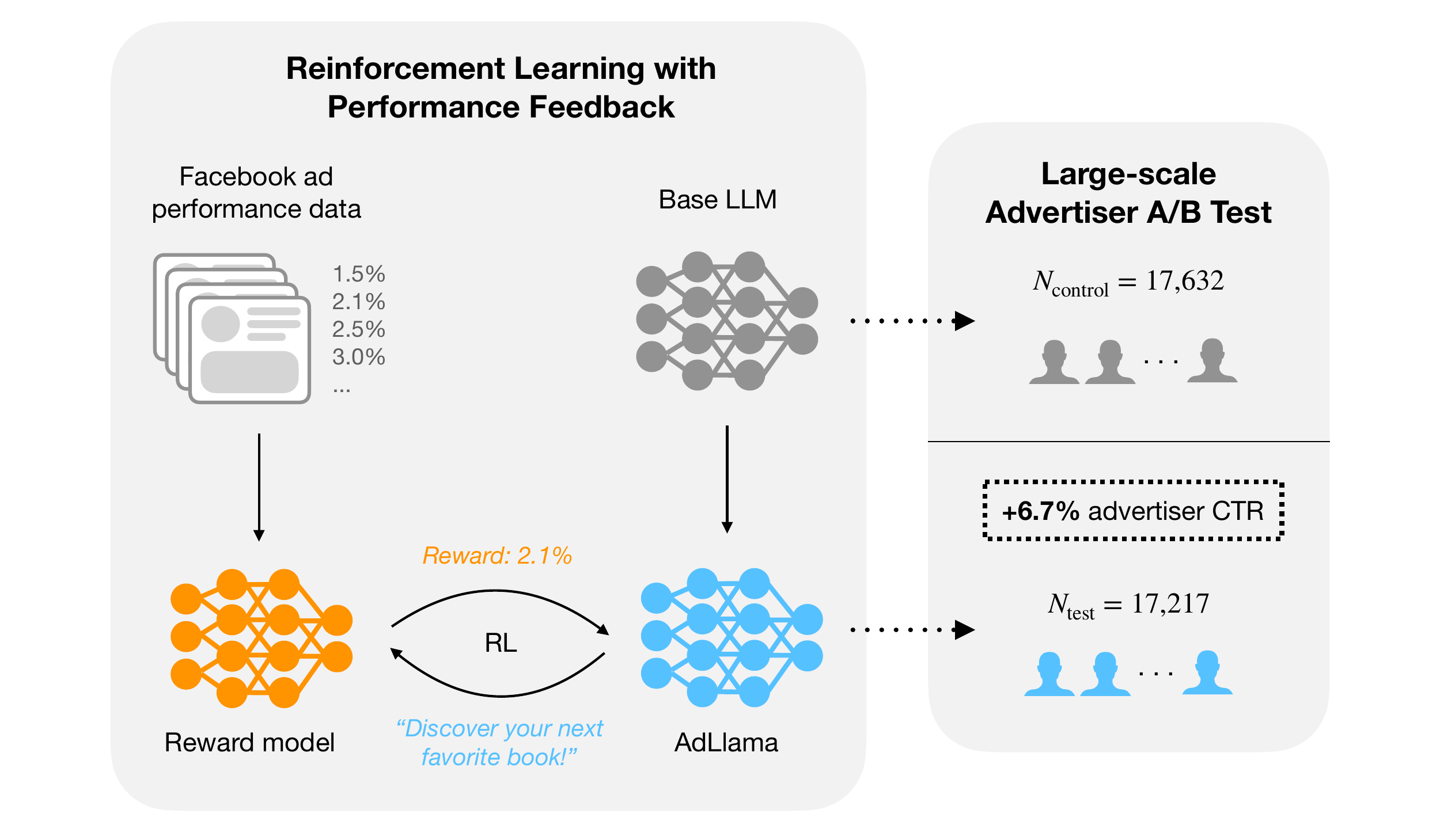}
	\caption{\textbf{Overview of our contributions.} Our first contribution, illustrated in the left panel, is RLPF, a reinforcement learning (RL) approach to post-training an LLM based on an aggregate performance metric. We apply RLPF to a generative AI feature in Meta's Ads Manager that helps advertisers generate new ad text variations. To do this, we use Facebook ad performance data (i.e., click-through rates) to train a reward model, which can score the effectiveness of a piece of ad text. Subsequently, we align the LLM toward this reward model using RL, which involves iteratively generating and scoring ad text, illustrated in lower part of the figure. The resulting LLM is called ``AdLlama.'' Our second contribution, illustrated in the right panel, are the results of a large-scale advertiser A/B test, encompassing approximately 35,000 advertisers and 640,000 ad variations, which showed that advertisers who received AdLlama achieved 6.7\% higher advertiser-level click-through rates (CTR). To our knowledge, this study is the \emph{largest reported so far} that investigates the use of generative AI in an ecologically valid setting.}
\label{fig:overview}
\end{figure*}
\else
\begin{figure*}[!ht] %
	\centering
	\includegraphics[width=\textwidth]{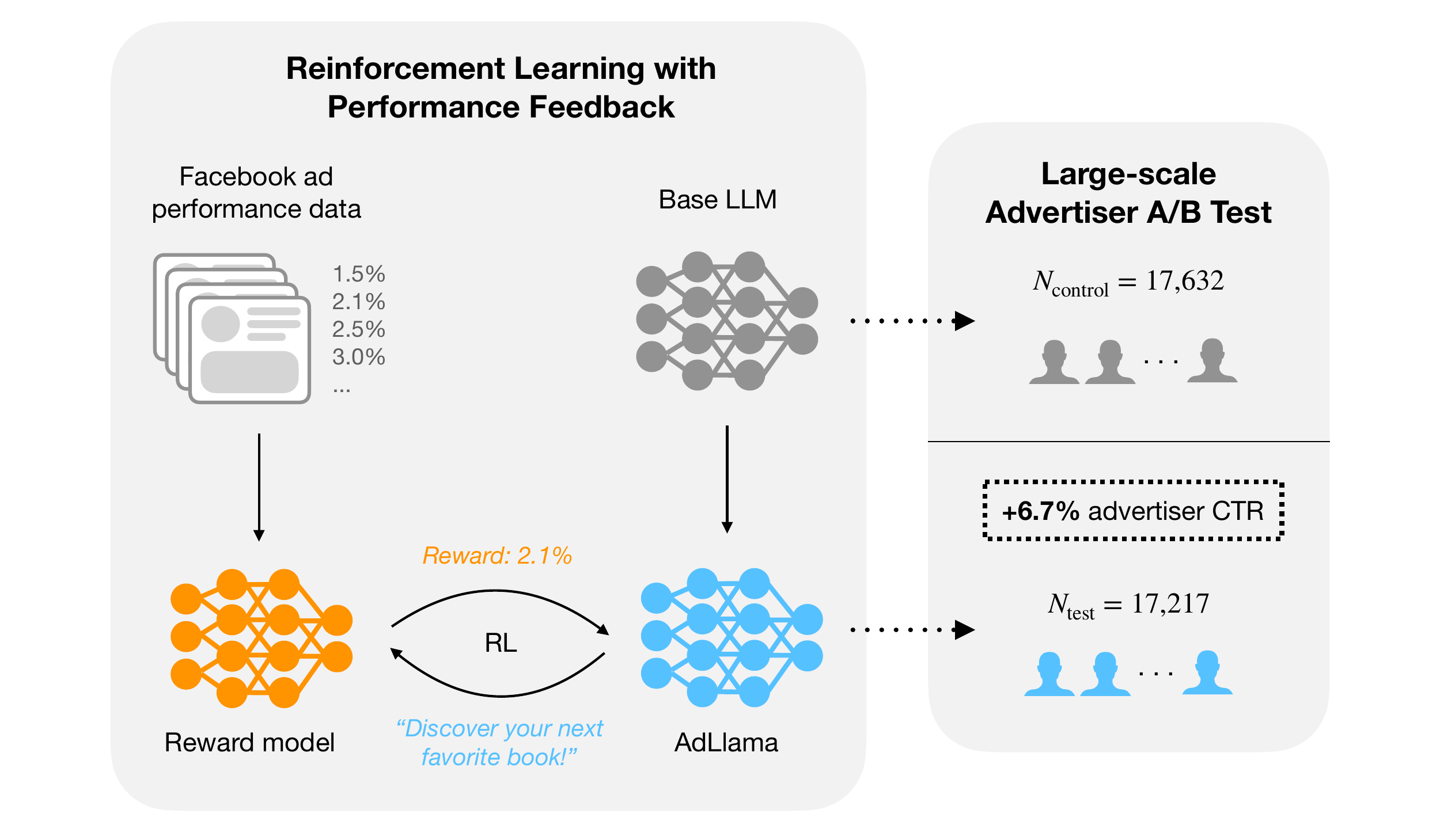}
	\caption{\textbf{Overview of our contributions.} Our first contribution, illustrated in the left panel, is RLPF, a reinforcement learning (RL) approach to post-training an LLM based on an aggregate performance metric. We apply RLPF to a generative AI feature in Meta's Ads Manager that helps advertisers generate new ad text variations. To do this, we use Facebook ad performance data (i.e., click-through rates) to train a reward model, which can score the effectiveness of a piece of ad text. Subsequently, we align the LLM toward this reward model using RL, which involves iteratively generating and scoring ad text, illustrated in lower part of the figure. The resulting LLM is called ``AdLlama.'' Our second contribution, illustrated in the right panel, are the results of a large-scale advertiser A/B test, encompassing approximately 35,000 advertisers and 640,000 ad variations, which showed that advertisers who received AdLlama achieved 6.7\% higher advertiser-level click-through rates (CTR). To our knowledge, this study is the \emph{largest reported so far} that investigates the use of generative AI in an ecologically valid setting.}
\label{fig:overview}
\end{figure*}
\fi

In this paper, we take the next step and describe our efforts in using reinforcement learning to improve the Text Generation LLM, with the measurable goal of writing \emph{more engaging} ad text. We do this by directly using performance (i.e., click-through rates) as a reward signal. We call the approach \emph{reinforcement learning with performance feedback} (RLPF). By viewing each user's behavior (i.e., click or no click) on each ad impression as a small piece of human feedback, RLPF can be thought of as an extension of the well-known RLHF technique \cite{ziegler2019fine,ouyang2022training}, except that each piece of ad text is being ``rated'' by \emph{thousands of humans} (i.e., Facebook users who see an ad impression) via click or no click signals. In that sense, one can view RLPF as falling somewhere in between traditional RLHF and the recently explored direction of RL with verifiable rewards \cite{lambert2024t}.

We report on the results of a large-scale online experiment (i.e., A/B test) conducted on Facebook that compared the effect of providing advertisers with the RLPF-trained model versus the naive SFT-based imitation model. Our experiment and analysis encompass approximately 35{,}000 advertisers and 640{,}000 ad variations, observed over 10 weeks. The results indicate that when equipped with the RLPF model, advertisers achieved a click-through rate (CTR) increase of \emph{6.7\%} compared to the naive imitation model. In addition, the number of ad variations created by the advertiser increased by 18.5\%, suggesting that advertisers were more willing to adopt AI suggestions when they came from AdLlama.

The implications of this result are significant. First, this improvement highlights the effectiveness of reinforcement learning as a methodology for metric-driven post-training of LLMs, especially in business use cases. Second, it offers a valuable data point by quantifying the benefits of RL-based post-training of LLMs in the domain of online advertising, contributing an important insight in the broader ongoing effort to understand the implications of generative AI. 

To our knowledge, is the \emph{largest study to date} examining the use of generative AI in a real-world, ecologically valid context. Previous analyses often relied on smaller-scale experiments involving a few thousand participants \cite{dell2023navigating,peng2023impact,ni2024generative,cui2024effects,angelopouloscausal} or controlled laboratory settings using crowd workers \cite{noy2023experimental,choi2023ai,schwarcz2025ai,chen2024large,ju2025collaborating}. These limitations largely reflect the high computational costs of training and deploying such models, as well as the scarcity of opportunities for large-scale field deployment.

\ifistwocolumn
\begin{figure*}[!ht]
	\centering
	\includegraphics[width=0.9\textwidth]{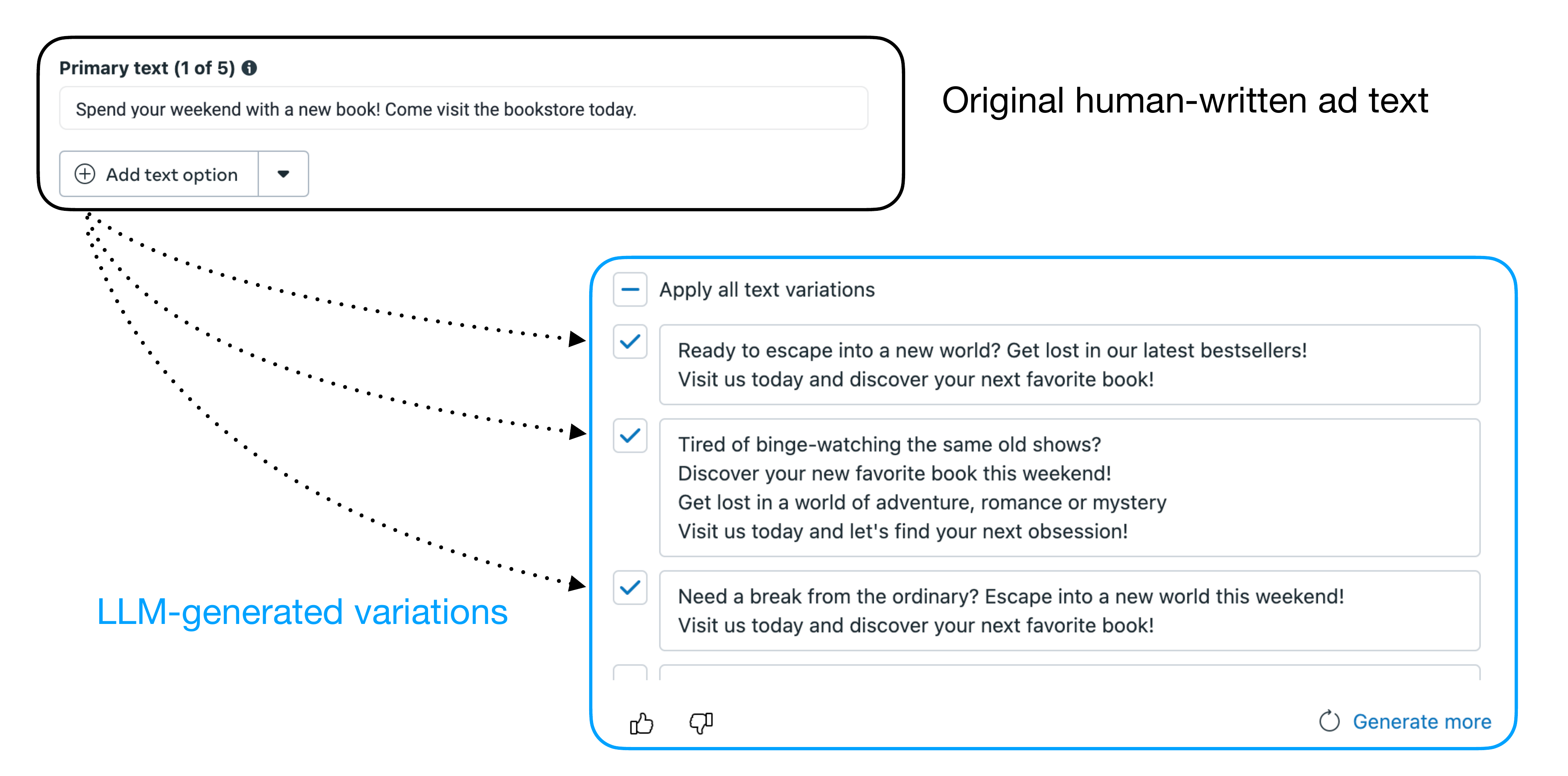} 
	\caption{\textbf{The user interface of the Meta Text Generation product.} Here we show an example of how the Text Generation product functions during the ad creation process. The advertiser provides an original version of the ad text (``\emph{Spend your weekend with a new book! Come visit the bookstore today.}''), which serves as input to Text Generation. The underlying LLM then produces a set of ad text variations, allowing the advertiser to pick and choose the ones they prefer to use. The advertiser can also opt to generate additional variants via the ``Generate more'' button.}
\label{fig:example}
\end{figure*}
\else
\begin{figure*}[!ht]
	\centering
	\includegraphics[width=\textwidth]{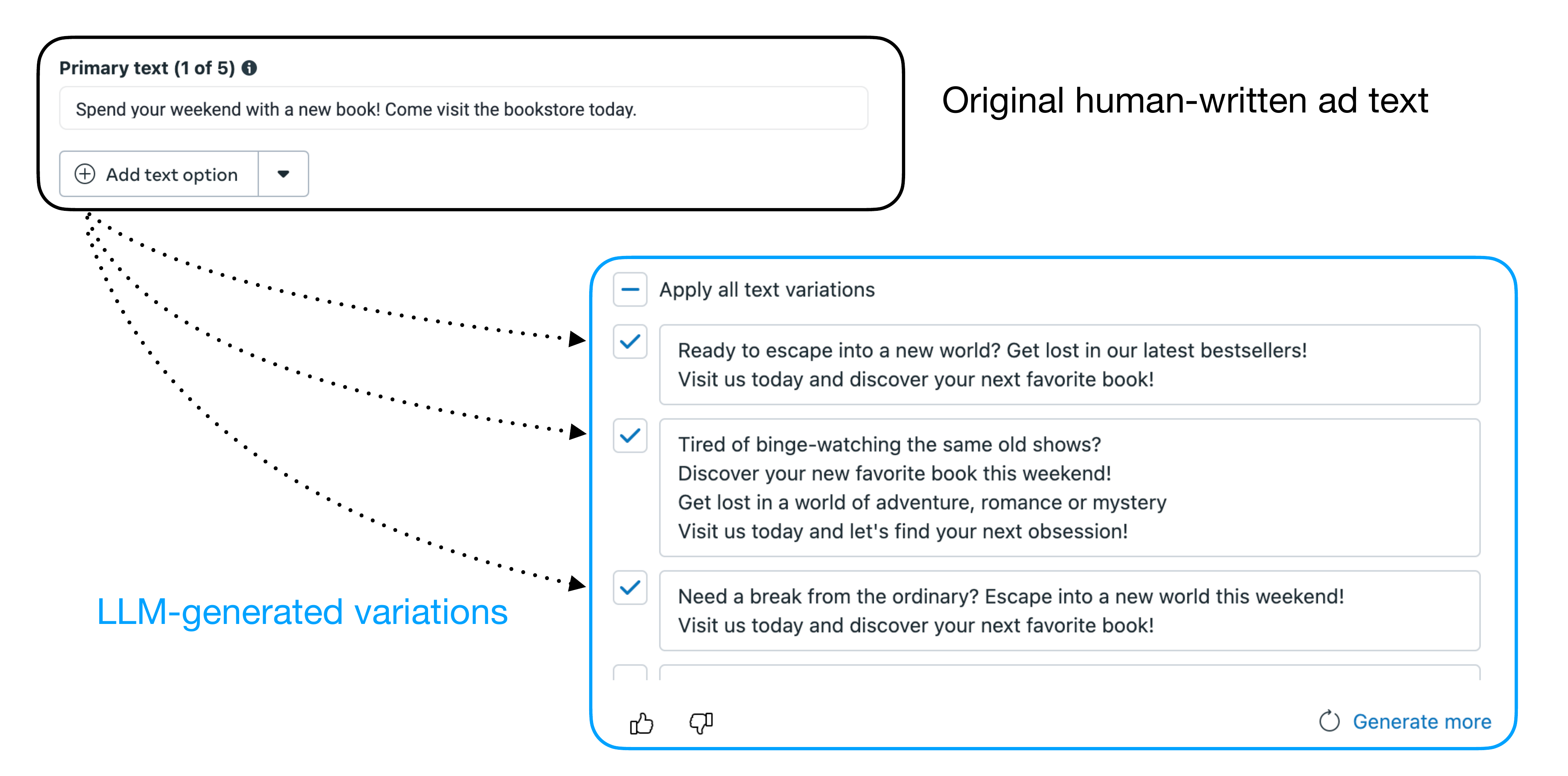} 
	\caption{\textbf{The user interface of the Meta Text Generation product.} Here we show an example of how the Text Generation product functions during the ad creation process. The advertiser provides an original version of the ad text (``\emph{Spend your weekend with a new book! Come visit the bookstore today.}''), which serves as input to Text Generation. The underlying LLM then produces a set of ad text variations, allowing the advertiser to pick and choose the ones they prefer to use. The advertiser can also opt to generate additional variants via the ``Generate more'' button.}
\label{fig:example}
\end{figure*}
\fi

\ifistwocolumn
\section*{Meta's Text Generation Product}
\else
\section{Meta's Text Generation Product}
\fi
The Text Generation product \cite{textgen1} is a generative AI feature within Meta Ads Manager that allows advertisers to experiment with multiple versions of their ad text. The feature works by taking an advertiser's original ad text as input and then using an underlying LLM to suggest new variations, which may, for example, emphasize key selling points or add additional creative messaging.

\ifistwocolumn
\subsection*{User Interface}
\else
\subsection{User Interface}
\fi
The user interface of the Text Generation product is illustrated in ~Figure~\ref{fig:example}. The flow is as follows. Advertisers first input an original ad text. The underlying LLM then generates and displays multiple text variations. Advertisers can then select the variations they want to use, edit them directly in the text box, or even add their own custom variations via the ``Add text option'' button \cite{textgen2}. In Figure~\ref{fig:example}, we show an example where the advertiser has selected a total of \emph{four} text variations, the original ad text along with three AI-generated variations. The option to continue generating additional text variations for consideration is available  via the ``Generate more'' button, but the advertiser is limited to selecting a maximum of five variations for delivery to users.

We note that although the Text Generation interface appears during the ad creation process, advertisers are \emph{never required} to select AI-written ads. Some other possible actions that the advertiser may opt to take are: (1) ignore the AI's suggestions completely and continue with the original text, (2) ignore the suggestions and instead supply multiple human-written text variations, (3) edit the AI-written ads before selecting them, (4) use the AI-written suggestions as inspiration for additional human-written variations, or (5) make AI-inspired edits to the original human-written variation after seeing the AI-generated variants. Therefore, the LLM can play an nuanced role in shaping ad text, regardless of whether the advertiser ultimately selects the precise wording that was generated.

\ifistwocolumn
\subsection*{Naive Imitation Models}
\else
\subsection{Naive Imitation Models}
\fi
\looseness-1 The original (naive) version of Text Generation LLM was released to advertisers in November 2023. This LLM is based on Meta's open-source foundation language model, the 7 billion (7B) parameter version of Llama 2 Chat \cite{touvron2023llama2}. As mentioned previously, the Llama model was post-trained to imitate a set of curated ads using SFT. We refer to this model as ``Imitation LLM v1.'' We refer to an improved version of the imitation model that uses higher quality data as ``Imitation LLM v2.'' The main difference between Imitation LLM v1 and v2 is that while the v1 training dataset is fully based on synthetic generations from a larger LLM, the v2 data additionally included human-written (i.e., contractor-written) examples. These training examples, whether synthetic or human-written, were curated by asking either the LLM or human to rewrite existing ads using specific instructions, such as ``paraphrase and shorten,'' ``make clear,'' ``make actionable,'' ``empathize,'' ``pose as a question,'' or ``focus selling point.'' We detail the training process of both Imitation LLM models in Section \ref{app:imitation_llm} of the supplementary materials.

The work presented in this paper is subsequent to the initial release of the Text Generation product. Our goal in this paper is to improve the initial imitation-based Text Generation LLM, focusing on \emph{quantifiably improving advertiser performance} relative to the original model, as measured in terms of click-through rate (CTR). We tackle this problem using a new idea, reinforcement learning on aggregate performance feedback signals.

\begin{figure*}[ht!] %
	\centering
	\includegraphics[width=\textwidth]{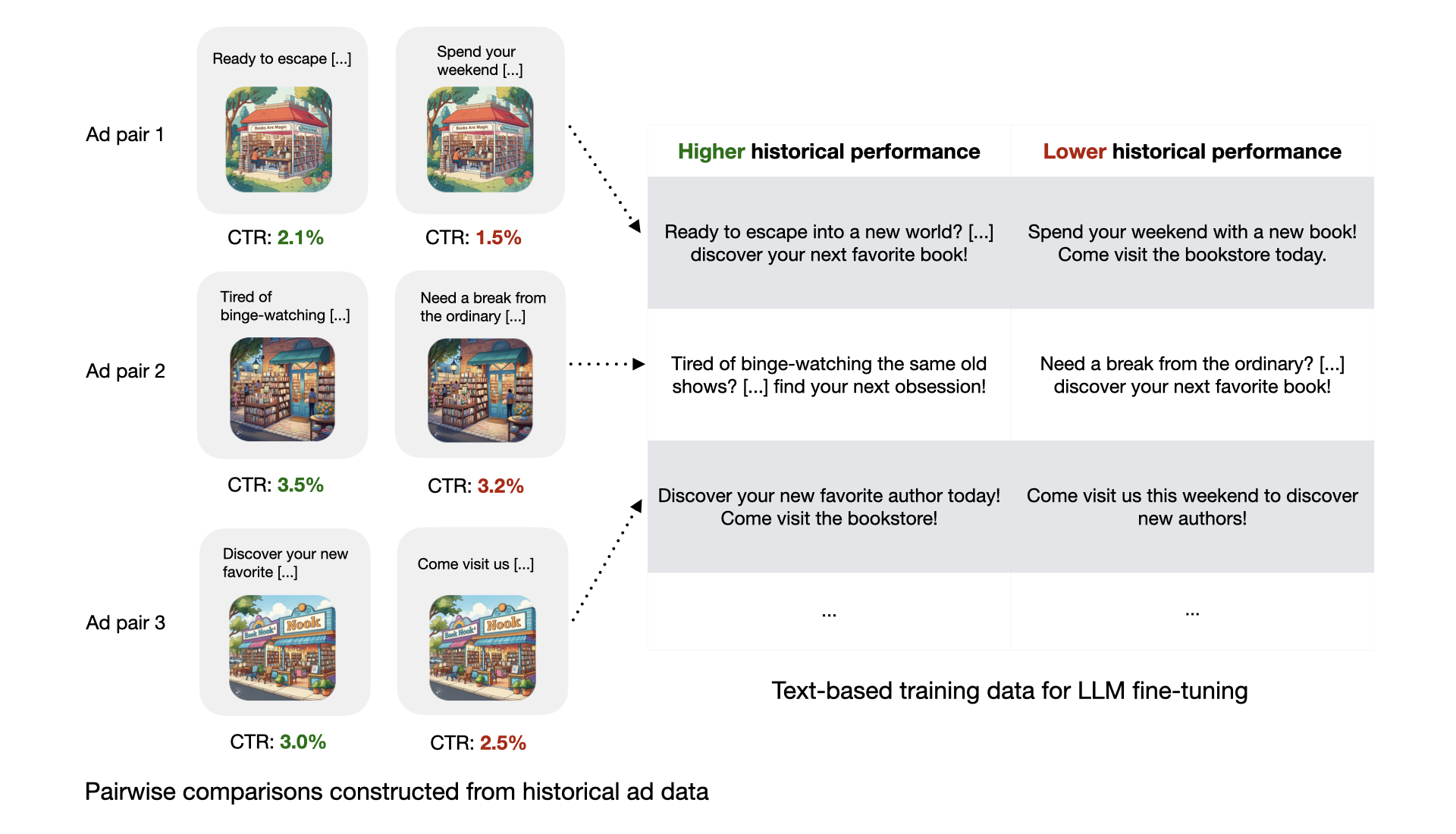}
	\caption{\textbf{Pairwise training data from historical multitext performance outcomes.} Based on \emph{multitext data}, data where advertisers change only the ad text while keeping all other ad components fixed, we are able to construct a pairwise training data point that distinguishes between text variations that are preferred and not preferred, as measured by CTR. On the left, we show historical ad pairs that have different text, but the same image. On the right, we show that each ad pair contributes to a single preference row for our training data.}
\label{fig:pairwise_training}
\end{figure*}

\ifistwocolumn
\section*{Methods}
\else
\section{Methods}
\fi
In this section, we first describe the methodology (including data preparation, reward model design, and reinforcement learning) for training the new version of the Text Generation LLM, which we call ``AdLlama.'' We then discuss the design of the experiment (i.e., A/B test) used in quantifying the performance improvement of the new model compared to the naive imitation model.

\ifistwocolumn
\subsection*{Reinforcement Learning with Performance Feedback (RLPF)}
\else
\subsection{Reinforcement Learning with Performance Feedback (RLPF)}
\fi
Although pre-trained LLMs have absorbed vast amounts of knowledge, they are typically not ready for widespread use. A crucial step before deployment to users is to \emph{align} the model for use on downstream tasks, which may include summarization \cite{stiennon2020learning}, answering questions \cite{nakano2021webgpt}, engaging in dialogue \cite{touvron2023llama}, or following a wide range of instructions from users \cite{ouyang2022training}. The primary approach for LLM alignment involves collecting \emph{preference} data from human labelers, who compare two responses and indicate which is better. The model is then fine-tuned on the preference data, encouraging it to generate responses more similar to those preferred by humans \cite{ziegler2019fine,ouyang2022training}. This process is known as \emph{reinforcement learning with human feedback} (RLHF). Because the ``quality'' of LLM responses for many standard LLM tasks (e.g., open-ended dialogue, creative writing) is subjective and hard to quantify in nature, training on human preferences is often the closest that one can come to a well-defined optimization objective.

Our key insight is that, unlike the above LLM use-cases, the task of \emph{crafting effective ad text} can be clearly associated with a \emph{quantifiable and measurable} objective: the CTR\footnote{The CTR is defined as the ratio of number of ad clicks to the total number of ad impressions (i.e., views). We note that CTR is typically considered a proxy metric for true goal of \emph{conversion rate} (the number of conversions, or purchases, per impression). However, because conversion rate is a far noisier metric to work with due to the sparsity of conversions, we resort to CTR.} of the ad. Note that such a setup holds anytime there is a concrete performance metric and exists beyond online advertising (e.g., e-commerce, AI customer support agents, ed tech). We propose the following \emph{general} approach, which can be thought of as a metric-driven extension of RLHF:
\begin{enumerate}
    \item First, using aggregate performance metrics, we train a performance \emph{reward model}, a model that can assign a reward (i.e., a score) to a piece of text, where more performant text are given higher scores.
    \item Subsequently, we use the reward model as an interactive environment to perform RL fine-tuning, where the goal is to fine-tune the LLM to be more likely to generate high-reward text.
\end{enumerate}

We now describe how we train a CTR-based performance reward model. Even prior to the launch of the Text Generation product, there existed a common practice among advertisers of testing multiple (human-written) text variants for a single ad using one of Meta's advertiser tools called ``Multiple Text Optimization'' \cite{multitext}. This practice enables us to observe historical ad data where, except for the text, all other components of the ad---such as the image, title, and targeting criteria---remain constant. We refer to this as \emph{multitext} data. 

From the multitext data, we are able to construct \emph{preference pairs}, where the higher CTR text is marked as ``more preferred'' and the lower CTR is marked as ``less preferred.'' See Figure \ref{fig:pairwise_training} for an illustration of this process. We call this the \emph{pairwise} dataset, which supports the standard Bradley-Terry preference-based approach to reward model training \cite{christiano2017deep,ziegler2019fine}. We also considered a more naive reward modeling approach via a \emph{pointwise} dataset, where each row is simply the ad text and its resulting CTR. We can then use a standard supervised learning  to train a reward model directly using CTR as labels. However, we found that the pointwise reward model was less capable of discerning ordering (or ranks) between similar pieces of ad text, which is ultimately more important than purely predicting CTR \cite{mandi2022decision,tan2024asymptotically}. Further details on the data curation and reward model training are in Sections \ref{app:training_data} and \ref{app:RLPF} of the supplementary materials.

\ifistwocolumn
\begin{figure*}[t!] %
	\centering
	\includegraphics[width=0.9\textwidth]{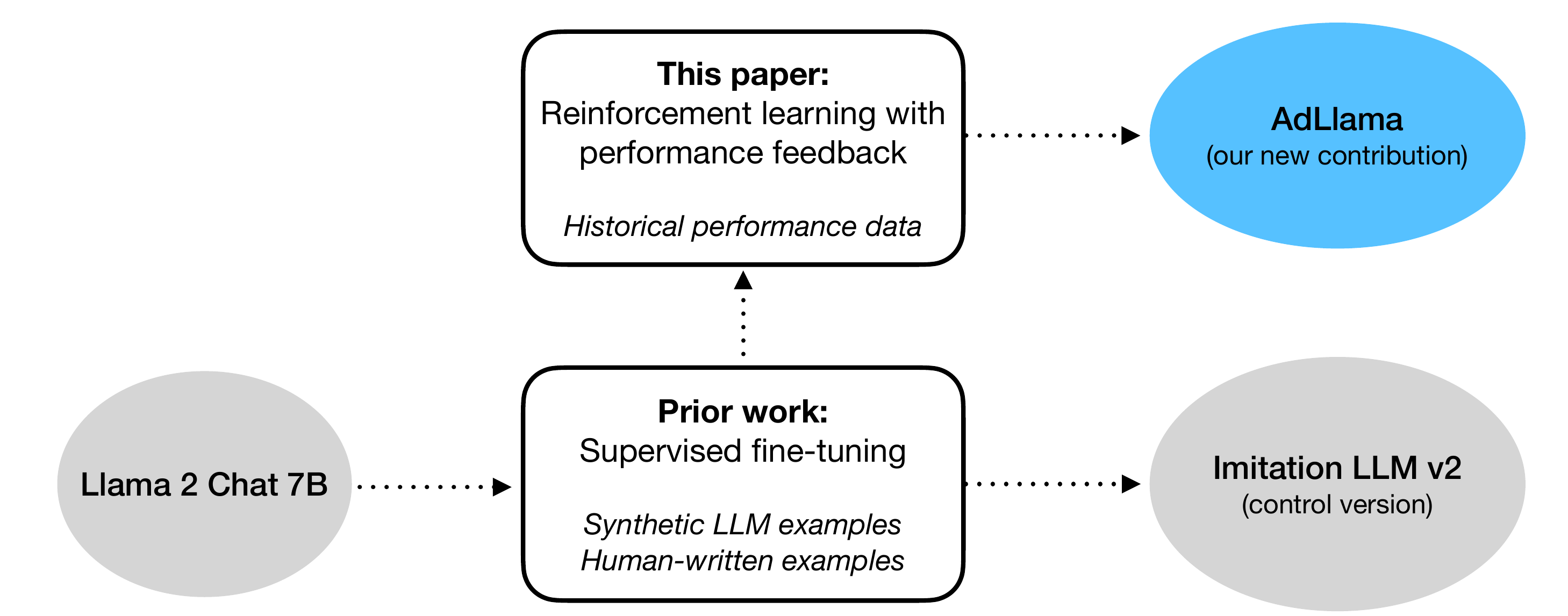}
	\caption{\textbf{AdLlama versus Imitation LLM v2.} Both AdLlama and Imitation LLM v2 originate from a base 7B Llama 2 Chat model. The difference is that AdLlama is further trained via RLPF and historical ad performance data, while Imitation LLM v2 is only trained using SFT to imitate a set of curated examples (which includes both LLM-generated synthetic examples and human-written examples).}
\label{fig:model_comparison}
\end{figure*}
\else
\begin{figure*}[t!] %
	\centering
	\includegraphics[width=\textwidth]{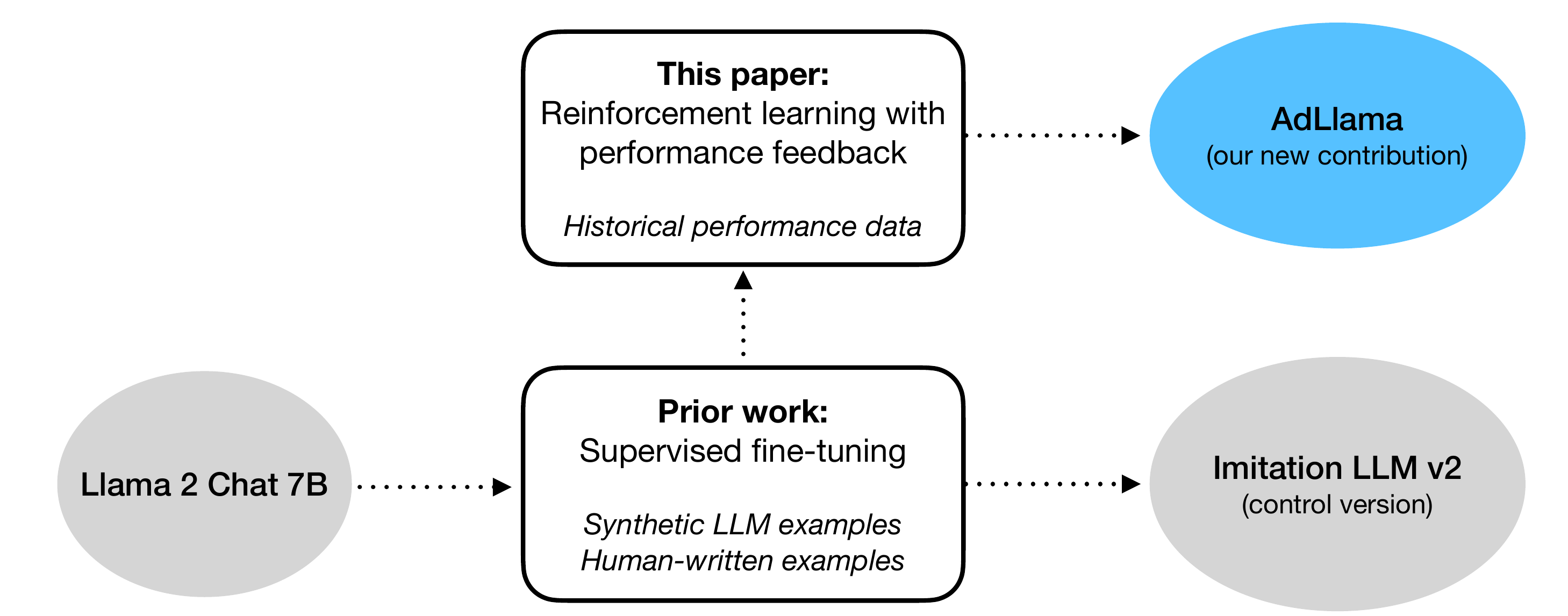}
	\caption{\textbf{AdLlama versus Imitation LLM v2.} Both AdLlama and Imitation LLM v2 originate from a base 7B Llama 2 Chat model. The difference is that AdLlama is further trained via RLPF and historical ad performance data, while Imitation LLM v2 is only trained using SFT to imitate a set of curated examples (includes both LLM-generated synthetic examples and human-written examples).}
\label{fig:model_comparison}
\end{figure*}
\fi

Given a trained reward model $r_\theta$, we use the proximal policy optimization (PPO) algorithm \cite{ouyang2022training,schulman2017proximal} to align the LLM with high-performance ad text. We added a length penalty to counteract the tendency for the model to generate ad text that is too long, leading to the following PPO optimization formulation:
\ifistwocolumn
    \begin{align*}
    \max_{\pi_\phi} \, \mathbb{E}_{x \sim \mathcal D_\textnormal{LLM}, y\sim \pi_\phi(y|x)} \bigl[ r_\theta(x, y)  &- \beta \, \textnormal{KL} [ \pi_\phi( \cdot \mid x), \pi_\textnormal{ref}( \cdot \mid x)  ] \\
    &- \alpha \, \textnormal{length}(y) \bigr],
    \end{align*}
\else
    \[
    \max_{\pi_\phi} \, \mathbb{E}_{x \sim \mathcal D_\textnormal{LLM}, y\sim \pi_\phi(y|x)} \bigl[ r_\theta(x, y)  - \beta \, \textnormal{KL} [ \pi_\phi( \cdot \mid x), \pi_\textnormal{ref}( \cdot \mid x)  ] - \alpha \, \textnormal{length}(y) \bigr],
    \]
\fi
where $\pi_\phi$ is the LLM, $\mathcal D_\textnormal{LLM}$ is the pointwise dataset used for LLM post-training, $\textnormal{KL}(p,q)$ is the Kullback-Leibler divergence between two probability distributions $p$ and $q$, $\textnormal{length}(y)$ is the number of tokens in $y$, and $\beta$ and $\alpha$ are weight parameters.

Full details of the approach are given in Section \ref{app:RLPF} of the supplementary materials. 
We used the RLPF technique to improve Imitation LLM v2, which, like Imitation LLM v1, is based on the 7B version of Llama 2 Chat \cite{touvron2023llama}. We refer to our RLPF-based ad text generation model as ``AdLlama.'' An illustrative comparison of the models is given in Figure \ref{fig:model_comparison}, where we emphasize both the difference in training method (RLPF versus SFT) and the training data (historical ad performance versus curated examples).

\ifistwocolumn
\subsection*{Experiment Design}
\else
\subsection{Experiment Design}
\fi

\ifistwocolumn
\begin{figure*}[t!]
	\centering
	\includegraphics[width=0.9\textwidth]{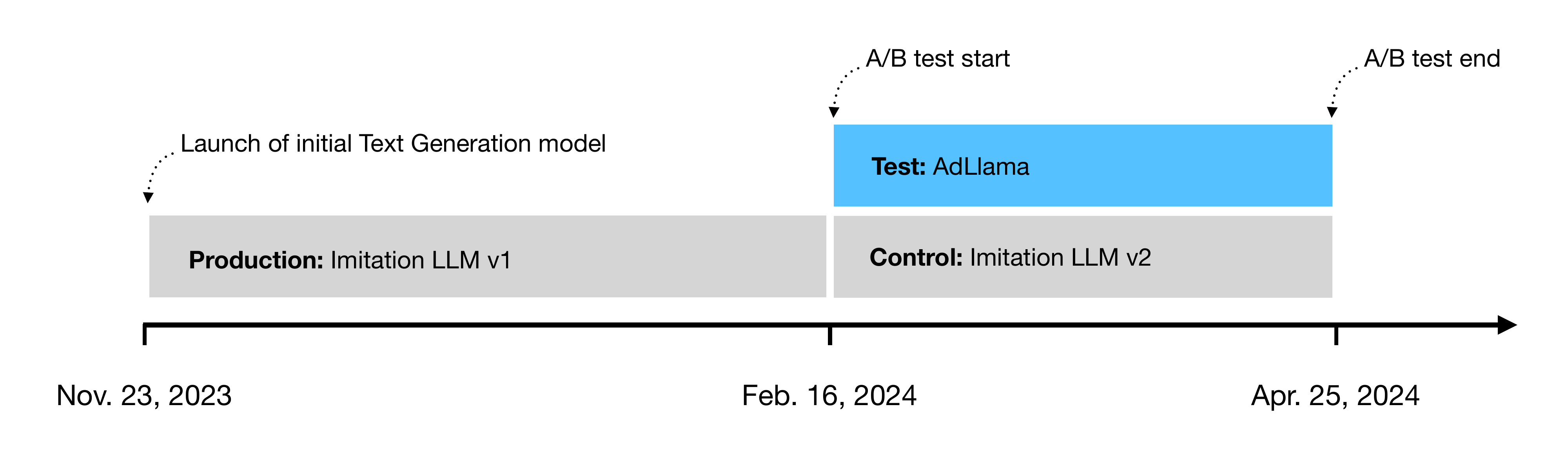}
	\caption{\textbf{A/B test timeline.} Our A/B test ran from February 16, 2024 until April 25, 2024. Prior to that, Imitation LLM v1 was launched on November 23, 2023 as the initial version of the Text Generation LLM.}
\label{fig:timeline}
\end{figure*}
\else
\begin{figure*}[t!]
	\centering
	\includegraphics[width=\textwidth]{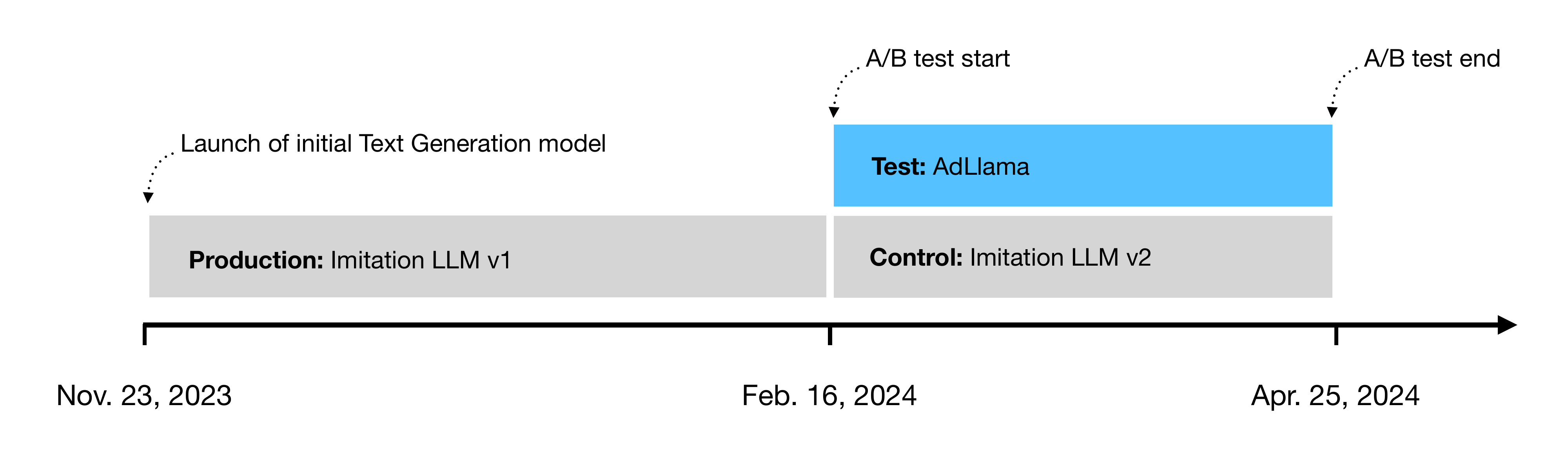}
	\caption{\textbf{A/B test timeline.} Our A/B test ran from February 16, 2024 until April 25, 2024. Prior to that, Imitation LLM v1 was launched on November 23, 2023 as the initial version of the Text Generation LLM.}
\label{fig:timeline}
\end{figure*}
\fi

\looseness-1 We conducted a large-scale A/B test (i.e., randomized control trial) to evaluate the impact of RLPF training on advertiser performance by comparing the AdLlama model against Imitation LLM v2.
The A/B test ran for 10 weeks, from February 16, 2024 until April 25, 2024, on $N=34{,}849$ advertisers based in the United States. We randomize at the advertiser level: each advertiser is randomly assigned either (1) the Imitation LLM v2 (``control'') or (2) the AdLlama LLM trained with RLPF (``test''). Figure \ref{fig:timeline} shows the timeline of the A/B test and how it relates to the launch of the initial Imitation LLM v1 model.

Our primary focus is on \emph{advertiser-level performance}, defined as performance aggregated over all \emph{direct-response}\footnote{There are two broad categories of ads, \emph{direct-response}  and \emph{brand/awareness}. \emph{Direct-response} ads are those that have a goal of optimizing for some form of engagement or immediate action from the user. Examples of direct-response advertising goals include clicks, purchases, app installs, sign ups, donations, or subscriptions. On the other hand, \emph{brand} ads aim to drive awareness simply through views of the ad (or video), and the goal is not to drive engagement from the user. Since the AdLlama model is designed to improve engagement, the relevant metric for evaluating AdLlama is CTR of direct-response ads.} ads created by the advertiser\footnote{A frequent practice of advertisers is to duplicate existing ads, some of which may have been originally created prior to the experiment start date. Such ads are excluded from our results, but duplicates of ads created \emph{during} the experimental period are included.} over the 10-week experimental period. We chose to examine advertiser-level performance because our goal is to understand how generative AI can improve advertiser return on investment. More specifically, our analysis centers on the following metrics: total engagement (clicks), total impressions (views), total number of ads created, and total number of ad variations created,\footnote{Recall from Figure \ref{fig:example} that in the Text Generation product, a single ``ad'' is associated with one or more ``variations'' where the body text is altered.} all of which are defined at the advertiser level across the 10-week experimental period on Facebook mobile feed.

We also log a number of advertiser covariates, which are listed in full in Section \ref{app:table_one} of the supplementary materials, along with detailed descriptions. Two notable covariates are: the advertiser's pre-experiment lifetime CTR across all Meta apps (``pre\_exp\_ctr'') and whether or not the advertiser is ``new,'' defined as whether the advertiser's pre-experiment ads have accumulated fewer than 1{,}000 impressions (``is\_new\_advertiser''). We also include the advertiser's ad creation behavior during the period after Imitation LLM v1 was released, but before the experiment started (``nov\_feb\_ad\_cnt'', ``nov\_feb\_variant\_cnt'', ``has\_created\_llm\_ad''). This period of time is highlighted in blue in Figure \ref{fig:timeline} and represents the initial few months of the Text Generation feature being available to advertisers. 

We also include other advertiser characteristics: vertical, expertise level, budget level, business account status, and account age. In addition to covariate details, we also present descriptive statistics and the balance of these covariates across conditions in Section \ref{app:table_one} of the supplementary materials (see Table \ref{table:covariates}).  We do not observe statistically significant imbalances between the two groups.

\ifistwocolumn
\section*{Main Results}
\else
\section{Main Results}
\fi

\ifistwocolumn
\subsection*{Advertiser Performance}
\else
\subsection{Advertiser Performance}
\fi
We aim to evaluate the effect of AdLlama and Imitation LLM v2 on advertiser-level CTRs. Our main regression specification is a log-binomial\footnote{Binomial regression is a natural candidate for modeling a CTR, since we are interested in clicks (successes) relative to impressions (total trials). Our use of a log link function leads to a relative risk rather than an odds ratio \cite{donoghoe2018logbin}. The relative risk is more desirable in our setting since it can be directly interpretable as a ratio of CTRs.} regression model, with the left-hand side representing the logarithm of the CTR of advertiser $i$:
\ifistwocolumn
\begin{equation}
\label{eq:main_binom}
\begin{aligned}
    \log \mathbb{E}(Y_i / n_i \mid X_i) &= \beta_0 + \beta_1 \cdot \textnormal{AdLlama}_i \\
    &\quad + \beta_2 \cdot \mathbf{1}_{\textnormal{existing}(i)} \cdot \log(\textnormal{CTR}^{\text{pre}}_i) \\
    &\quad + \beta_3 \cdot \mathbf{1}_{\textnormal{new}(i)} + \boldsymbol{\beta}_{4{:}10}^T Z_i \\
    &\quad + \theta_{\textnormal{vert}(i)} + \alpha_{\textnormal{budget}(i)} + \lambda_{\textnormal{expertise}(i)} \vphantom{\log(\textnormal{CTR}^{\text{pre}}_i)},
\end{aligned}
\end{equation}
\else
\begin{equation}
\label{eq:main_binom}
\begin{aligned}
    \log \mathbb{E}(Y_i / n_i \, | \, X_i) = \beta_0 &+ \beta_1 \cdot \textnormal{AdLlama}_i + \beta_2 \cdot \mathbf{1}_{\textnormal{existing}(i)} \cdot \log(\textnormal{CTR}^{\text{pre}}_i) + \beta_3 \cdot \mathbf{1}_{\textnormal{new}(i)}  \\
    &  + \boldsymbol{\beta}_{4:10}^T \, \, Z_i + \theta_{\textnormal{vert}(i)} + \alpha_{\textnormal{budget}(i)} + \lambda_{\textnormal{expertise}(i)},
\end{aligned}
\end{equation}
\fi
where $Y_i$ is the number of clicks (engagement) for the $i$-th advertiser, $n_i$ is the impression count, $X_i$ is the set of all covariates, $\textnormal{CTR}^{\text{pre}}_i$ is the pre-experiment CTR (``pre\_exp\_ctr''), $\textnormal{AdLlama}_i$ is a binary indicator for the AdLlama LLM treatment (which is randomly assigned), $\mathbf{1}_{\textnormal{existing}_i}$ is an indicator for ``$\textnormal{is\_new\_advertiser}$ equal to $0$,'' $\mathbf{1}_{\textnormal{new}_i}$ is an indicator for ``$\textnormal{is\_new\_advertiser}$ equal to $1$,'' $\theta_{\textnormal{vert}(i)}$ are vertical fixed effects, $\alpha_{\textnormal{budget}(i)}$ are budget category fixed effects, $\lambda_{\textnormal{expertise}(i)}$ are expertise category fixed effects, and $Z_i$ is the vector of remaining numerical covariates. Note that $\mathbb{E}(Y_i / n_i \, | \, X_i)$ is the CTR of advertiser $i$.

\begin{table*}[!t] \centering 
    \footnotesize
  \caption{Log-binomial regression results on advertiser-level (engagement, impressions), with the ``treatment'' variable being the indicator for using AdLlama. The variable ``log\_pre\_exp\_ctr\_existing''  refers to the product $\mathbf{1}_{\textnormal{existing}(i)} \cdot \log(\textnormal{CTR}^{\text{pre}}_i)$. Coefficients are reported on the link function scale (log scale); the $6.7\%$ improvement quoted in the main text computed by $6.7\% \approx \exp(0.0651)-1$. Fixed effect terms are omitted for brevity. All reported p-values are two-sided.} 
\vspace{10pt}
\label{table:log_binom} 
\begin{tabular}{@{\extracolsep{5pt}}lc} 
\\[-1.8ex]\hline 
\hline \\[-1.8ex] 
 & \multicolumn{1}{c}{\textit{Dependent variable:}} \\ 
\cline{2-2} 
\\[-1.8ex] & engagement/impressions \\ 
\hline \\[-1.8ex] 
 \textbf{treatment} & \textbf{0.0651}$^{**}$ \textbf{(0.0299)} \\ 
  log\_pre\_exp\_ctr\_existing & 0.5931$^{***}$ (0.0264) \\ 
  is\_new\_advertiser & $-$2.3469$^{***}$ (0.1569) \\ 
  pre\_exp\_ad\_cnt & 0.0024 (0.0028) \\ 
  pre\_exp\_impressions & $-$0.0001 (0.0001) \\ 
  pre\_exp\_engagement & 0.0036 (0.0049) \\ 
  account\_age\_yr & 0.0008 (0.0038) \\ 
  nov\_feb\_ad\_cnt & 0.0013 (0.0019) \\ 
  nov\_feb\_variant\_cnt & $-$0.0003 (0.0006) \\ 
  is\_business\_account & $-$0.0286 (0.0424) \\ 
  has\_created\_llm\_ad & $-$0.0109 (0.0345) \\ 
  Constant & $-$1.1233$^{***}$ (0.1506) \\ 
 \hline \\[-1.8ex] 
Observations & 34,849 \\ 
\hline 
\hline \\[-1.8ex] 
\textit{Note:} HC1 robust standard errors in parentheses.  & \multicolumn{1}{r}{$^{*}$p$<$0.1; $^{**}$p$<$0.05; $^{***}$p$<$0.01} \\ 
\end{tabular} 
\end{table*}

Because some advertisers in the experiment had no pre-experiment ad impressions (or an insufficient number of impressions for a reliable CTR calculation), it is not possible to naively include the pre-experiment CTR as a covariate for all advertisers. Instead, we devise the following strategy to properly account for pre-experiment CTR through the $\beta_2$ and $\beta_3$ terms of Equation \ref{eq:main_binom}. If the advertiser is an ``existing'' advertiser (i.e., has sufficient ad impressions), then we directly incorporate its pre-experiment CTR via the term $\beta_2 \cdot \mathbf{1}_{\textnormal{existing}(i)} \cdot \textnormal{log pre\_exp\_ctr}_i$. This can be interpreted as a baseline log CTR for that advertiser, which is then ``adjusted'' by the regression specification based on other covariates. New advertisers do not have a reliable pre-experiment CTR, so we estimate a baseline log CTR using the term $\beta_3 \cdot \mathbf{1}_{\textnormal{new}(i)}$. In Section \ref{app:loglinkinterp} of the supplementary materials, we show how our regression specification above leads to an intuitive interpretation.

Table \ref{table:log_binom} presents the findings from the log-binomial regression analysis. The results indicate that AdLlama provides a statistically significant \textbf{6.7\% increase} in advertiser-level CTR ($p = 0.0296$ and a standard error of 0.0299) when compared to the naive Imitation LLM v2. This roughly corresponds to an absolute increase in advertiser-level CTR  from 3.1\% to 3.3\%. Although the absolute increase appears modest at first glance, a 6.7\% relative increase in CTR represents a substantial improvement in an advertiser's return on investment for Facebook ads. Furthermore, on mature, highly-optimized ad platforms like Facebook, even small increases in CTR are typically difficult to achieve.

\ifistwocolumn
\subsubsection*{Robustness Checks}
\else
\subsubsection{Robustness Checks}
\fi
We provide several robustness checks to validate the findings above. Since we did not observe significant imbalances between the control and test groups, we used model-free results as additional supporting evidence (see Section \ref{app:model_free} of the supplementary materials). Further, in Section \ref{app:logistic_pois} of the supplementary materials, we tested various alternative CTR regression specifications, including quasi-binomial, logistic, Poisson, and quasi-Poisson regressions. These analyses yielded qualitatively similar effects. Finally, in Section \ref{app:separate} of the supplementary materials, we conducted separate linear regressions on clicks and impressions, providing statistical evidence that AdLlama increases the total number of clicks per advertiser, while it did not affect the total number of impressions delivered for the advertiser. This is further supporting evidence that CTR is increased under AdLlama.

\ifistwocolumn
\subsection*{Impact on Ad Variations Created}
\else
\subsection{Impact on Ad Variations Created}
\fi
We are also interested in how AdLlama and Imitation LLM v2 affect advertisers' usage of the Text Generation product. We consider two outcomes, (1) the number of \emph{ad variations} created during the experimental period and (2) the number of ads created during the experimental period. Recall that a single ``ad'' is associated with one or more ``variations'' (see Figure \ref{fig:example}). 

We use a linear regression with the same covariates as we did in the log-binomial regression of Equation \ref{eq:main_binom}, except without the log-transformation\footnote{We remove the log-transformation because there is no longer a log link function; see Section \ref{app:loglinkinterp} of the supplementary materials for further discussion.} of the pre-experimental CTR:
\ifistwocolumn
\begin{equation}
\label{eq:variants_regression}
\begin{aligned}
    \text{Outcome}_i &= \beta_0 + \beta_1 \cdot \textnormal{AdLlama}_i \\
    &\quad + \beta_2 \cdot \mathbf{1}_{\textnormal{existing}(i)} \cdot \textnormal{CTR}^{\text{pre}}_i \\
    &\quad + \beta_3 \cdot \mathbf{1}_{\textnormal{new}(i)} + \boldsymbol{\beta}_{4{:}10}^T Z_i \\
    &\quad + \theta_{\textnormal{vert}(i)} + \alpha_{\textnormal{budget}(i)} + \lambda_{\textnormal{expertise}(i)} + \epsilon_i \vphantom{\textnormal{CTR}^{\text{pre}}_i}.
\end{aligned}
\end{equation}
\else
\begin{equation}
\label{eq:variants_regression}
\begin{aligned}
    \text{Outcome}_i = \beta_0 &+ \beta_1 \cdot \textnormal{AdLlama}_i + \beta_2 \cdot \mathbf{1}_{\textnormal{existing}(i)} \cdot \textnormal{CTR}^{\text{pre}}_i + \beta_3 \cdot \mathbf{1}_{\textnormal{new}(i)}  \\
    &  + \boldsymbol{\beta}_{4:10}^T \, \, Z_i + \theta_{\textnormal{vert}(i)} + \alpha_{\textnormal{budget}(i)} + \lambda_{\textnormal{expertise}(i)} + \epsilon_i.
\end{aligned}
\end{equation}
\fi
Here, $\text{Outcome}_i$ refers to either advertiser $i$'s variation count or ad count and $\epsilon_i$ is the error term. 
\begin{table*}[!t] \centering 
\footnotesize
  \caption{Linear regression results on advertisers' ad creation behavior: ``variant\_cnt'' is the number of ad variations created, ``ad\_cnt'' is the number of ads created, and ``pre\_exp\_ctr\_existing''  refers to the term $\mathbf{1}_{\textnormal{existing}(i)} \cdot \textnormal{CTR}^{\text{pre}}_i$. All other notation remains consistent with Table \ref{table:log_binom}. We report HC1 robust standard errors. Fixed effect terms are omitted for brevity. All reported p-values are two-sided.} 
  \label{table:lin_variants} 
  \vspace{10pt}

\begin{tabular}{@{\extracolsep{5pt}}lcc} 
\\[-1.8ex]\hline 
\hline \\[-1.8ex] 
 & \multicolumn{2}{c}{\textit{Dependent variable:}} \\ 
\cline{2-3} 
\\[-1.8ex] & variant\_cnt & ad\_cnt \\ 
\\[-4.8ex] & & \\ 
\hline \\[-1.8ex] 
 \textbf{treatment} & \textbf{3.113}$^{***}$ \textbf{(0.528)} & \textbf{0.040 (0.189)} \\ 
  pre\_exp\_ctr\_existing & $-$5.915 (11.057) & $-$6.568$^{*}$ (3.549) \\ 
  pre\_exp\_ad\_cnt & 3.803$^{***}$ (0.689) & 1.564$^{***}$ (0.280) \\ 
  pre\_exp\_impressions & 0.009 (0.011) & 0.002 (0.004) \\ 
  pre\_exp\_engagement & $-$0.333 (0.254) & $-$0.123 (0.098) \\ 
  account\_age\_yr & $-$0.397$^{***}$ (0.083) & $-$0.197$^{***}$ (0.029) \\ 
  nov\_feb\_ad\_cnt & 0.750$^{*}$ (0.441) & 0.615$^{***}$ (0.170) \\ 
  nov\_feb\_variant\_cnt & 0.096 (0.137) & $-$0.055 (0.050) \\ 
  is\_new\_advertiser & 2.656$^{***}$ (0.776) & 0.991$^{***}$ (0.268) \\ 
  is\_business\_account & 3.405$^{***}$ (0.420) & 1.416$^{***}$ (0.148) \\ 
  has\_created\_llm\_ad & 2.078$^{**}$ (1.033) & 0.280 (0.399) \\ 
  Constant & 7.797$^{***}$ (2.035) & 3.831$^{***}$ (0.685) \\ 
 \hline \\[-1.8ex] 
Observations & 34,849 & 34,849 \\ 
R$^{2}$ & 0.091 & 0.116 \\ 
Adjusted R$^{2}$ & 0.091 & 0.115 \\ 
\hline 
\hline \\[-1.8ex] 
\textit{Note:} HC1 robust standard errors in parentheses. & \multicolumn{2}{r}{$^{*}$p$<$0.1; $^{**}$p$<$0.05; $^{***}$p$<$0.01} \\ 
\end{tabular} 
\end{table*}

Table \ref{table:lin_variants} reports the results of these regressions. We observe strong evidence that usage of the AdLlama LLM \emph{increases the number of ad variations created by the advertiser}, while the total number of ads created remained statistically the same. Specifically, we see that the number of ad variations increased by 3.1 ($p < 0.01$) from roughly 16.8 variations (Imitation LLM v2) to 19.9 (AdLlama). This is an 18.5\% increase when using the AdLlama LLM. This suggests that for each ad, advertisers were more willing to use the Text Generation product's suggestions when they came from AdLlama compared to Imitation LLM v2.

\ifistwocolumn
\section*{Discussion}
\else 
\section{Discussion}
\fi
Our work shows that RL with performance feedback can be used to train LLMs to generate ad text that resonates with advertisers and drives measurable engagement from users on Facebook. Specifically, our large-scale A/B test on Meta's Text Generation product shows that the RLPF-based model significantly increases advertiser-level CTRs, along with the number of ad text variations that advertisers were willing to employ. These results support the concept of anchoring the fine-tuning process in real-world, aggregate performance metrics, rather than relying solely on human raters' preference feedback or rule-based rewards.

\ifistwocolumn
\subsection*{Limitations}
\else 
\subsection{Limitations}
\fi
There are several limitations of our work that we now discuss. Our model was trained using offline historical performance data. Therefore, this is equivalent to a single round of \emph{offline} RL, where there is no real-time interaction with the environment. To further refine our model, we could incorporate the performance outcomes of LLM-generated ads in an iterative process. This approach would align more closely with \emph{online} RL, where the model continuously interacts (by taking actions) with the environment and adapts based on real-time feedback in the form of rewards and transitions \cite{sutton2018reinforcement}. Such a system would be more capable of adapting to new trends and perhaps even discovering new ones through \emph{exploration}, the process of experimenting with new actions (i.e., trying out new ad text formats unseen in the historical data).

Our current model primarily focuses on ad performance, but other factors are also important to consider. As an example, there may be a trade-off between generating ads that perform well and those that exhibit high creativity. Additionally, the model's ability to adhere to specific advertiser instructions, such as maintaining a particular tone, is another important consideration. Addressing these aspects would require a multi-objective optimization approach to balance various objectives effectively. Finally, our model currently does not take into account the human component of the Text Generation product: before an ad text variation can be delivered to users, the advertiser must explicitly select that variation for delivery. An alternative way to train the future iterations of the RLPF reward model is to weigh the CTR by the likelihood that the text is selected by advertisers.

Beyond individual ad performance, platform-level factors, such as the diversity of the ad inventory, are also important for a positive user experience. Future work should explore strategies that simultaneously consider these other factors, while also optimizing for performance.

\ifistwocolumn
\subsection*{Broader Implications}
\else 
\subsection{Broader Implications}
\fi
\looseness-1 Our findings contribute to the growing body of literature on understanding the impact of LLMs. By quantifying the benefits of RL-based post-training in online advertising, we provide a concrete data point that highlights the potential for these models to ingest relevant performance metrics and subsequently create real business impact. The ability to generate more engaging ad content not only improves existing advertisers' return on investment, but could also lower the barrier to entry for new and inexperienced advertisers (e.g., small businesses) by reducing the need for extensive marketing expertise and resources.

Our methodology is not limited to online advertising: the principles of RLPF can be adapted to other domains where aggregate performance metrics are available. By using performance data as a feedback mechanism, organizations can fine-tune LLMs to optimize for their desired outcomes. For example, the core methodology can easily be extended to closely related settings like personalized email campaigns or e-commerce product descriptions. RLPF can also be extended to settings with multiple rounds of interactive feedback, such as AI customer support agents, using metrics like resolution rates, satisfaction scores, or user response times.

There are also less obvious settings where RLPF could be applied. For example, in online learning platforms, student performance data (test scores and engagement metrics) could guide the generation of adaptive learning content, while for certain public awareness campaigns (e.g., vaccination, energy consumption), performance data could enable LLMs to rewrite communication materials to better resonate with their intended audience.

Our work only takes the first step in demonstrating the potential of RL augmented with aggregate performance feedback. We believe this is a promising and generalizable approach that bridges the gap between highly capable language models and tangible outcomes.

\section*{Acknowledgments}
We gratefully acknowledge our close partnership on this project with the Monetization GenAI and Creative \& Guidance teams at Meta:  Yide Zhao, Yair Levi, Clare Zhang, Steven Barnett, Shenghong Wang, Meilei Jiang, Jerry Pan, Sanjian Chen, Shenxiu Liu, Zhonghua Qu, Xueting Yan, and Arghya Paul.

\ifistwocolumn
\else
\section*{Author Contributions}
A.N. and D.J. created reward model datasets and trained the reward model. D.J. and A.N. then post-trained the Text Generation LLM using RLPF, which resulted in the AdLlama model described in this paper. Y.C. curated the source datasets for both the reward model and Text Generation LLM. Y.C. and Y.B. managed the A/B testing process. D.J. drafted the initial version of the paper; all authors revised and reviewed the paper. Y.B. and Z.Z. advised on all aspects of the project.

\section*{Author Affiliations}
All authors are either current or former employees of Meta. D.J., A.N., Y.C., and Y.B. are currently employed by Meta, while Z.Z. is a former employee. The work described in this paper by Z.Z. was conducted during Z.Z.'s employment at Meta.
\fi

\ifistwocolumn
\else
\clearpage %
\fi
\bibliography{refs} %

\appendix
\counterwithin{figure}{subsection}
\counterwithin{table}{subsection}

\ifistwocolumn
\clearpage
\section*{Supporting Information Appendix (SI)}
\else
\section{Supplementary Materials}
\fi

\setcounter{figure}{0}                 %
\setcounter{table}{0}

\subsection{Training Data}
\label{app:training_data}

\subsubsection{Reward Model}
We obtained our reward model (RM) training data from the Multiple Text Options feature in Meta Ads Manager \cite{multitext}. This feature enables advertisers to manually submit various versions of body text, title text, and description text. Multiple Text Options then automatically tests and optimizes to find the ad variation that most effectively engages audiences. This feature depends entirely on \emph{manually-written} text variations and therefore can be seen as a precursor to the AI-driven Text Generation feature (the focus of this paper).

Recall that the Text Generation feature targets the generation of \emph{ad body text}. However, the CTR of an ad is not determined by its body text alone, but depends on multiple other factors, including the ad image, the ad's targeting criteria, or the ad's vertical (e.g., engagement on real estate ads is expected to be drastically different from restaurant ads). Thus, Multiple Text Options provided us with an important type of data: CTR data for ads that are \emph{identical across all dimensions except for the ad body text}. This allows us to attribute the difference in CTR directly to the ad body text.

We filtered this data to include ad variations written in English with body text between 100 and 1000 characters (we excluded ads that were too short or too long) with at least 2{,}000 impressions on Facebook. From these data, we created preference pairs of ad variations where the two individual variations only differ in the ad body text, as previously illustrated in Figure \ref{fig:pairwise_training}. Our final RM training dataset included approximately 7 million preference pairs.

\subsubsection{Language Model}
The training data used for language model post-training (i.e., after the RM is trained) is sourced in the same way as described above, with the only exception being that preference pairs are not constructed. We refer to this as the ``pointwise'' version of the dataset (as opposed to ``pairwise''). The pointwise dataset included approximately 5.5 million human-written text variation examples.

\subsection{Reinforcement Learning with Performance Feedback}
\label{app:RLPF}
We adapt the RLHF paradigm by replacing pairwise human feedback with performance feedback. 
This change allows us to align the LLM not just with the preferences of a single human annotator (whose preferences might be noisy and not necessarily representative of the advertiser's goals), but the ad's CTR, a real-world performance metric that summarizes how the ad interacts with both the delivery system and Facebook users. Since the CTR is computed using all impressions of the ad, it becomes a less noisy estimate than a label from a handful of human annotators. The RLPF training pipeline has two main components: (1) reward model (RM) training and (2) post-training the LLM.

For RM training, we use the pairwise CTR preference dataset described above in Section \ref{app:training_data} of the supplementary materials, where preference pairs are decided by comparing CTRs over thousands of impressions (or more). We denote each row of data with a prompt $x$ and a preference pair $(y_w, y_l)$. Following prior work on RLHF \cite{christiano2017deep,ziegler2019fine}, the RM is trained with an underlying Bradley-Terry preference assumption, namely that the probability of ad text $y_1$ being higher performing than ad text $y_2$ is given by
\ifistwocolumn
\begin{align*}
\mathbb{P}(y_1 \succ y_2 \mid x) &= \frac{\exp(r_\theta(x, y_1))}{\exp(r_\theta(x, y_1)) + \exp(r_\theta(x, y_2))} \\
&= \frac{1}{1 + \exp[-(r_\theta(x, y_1) - r_\theta(x, y_2))]}\\
&= \sigma(r_\theta(x, y_1) - r_\theta(x, y_2)),
\end{align*}
\else
\begin{align*}
\mathbb{P}(y_1 \succ y_2 \mid x) &= \frac{\exp(r_\theta(x, y_1))}{\exp(r_\theta(x, y_1)) + \exp(r_\theta(x, y_2))} \\
&= \frac{1}{1 + \exp[-(r_\theta(x, y_1) - r_\theta(x, y_2))]} = \sigma(r_\theta(x, y_1) - r_\theta(x, y_2)),
\end{align*}
\fi
where $r_\theta(x, y)$ is the reward model with parameters $\theta$ and $\sigma$ is the logistic function. The reward $r_\theta(x,y)$ represents the ``strength'' of response $y$ given prompt $x$. Therefore, the Bradley-Terry model relates the preference to the relative strengths of two pieces of ad text. To fit the parameters $\theta$, we use maximum likelihood, arriving at the negative log-likelihood loss function:
\[
\mathcal L_{\textnormal{RM}}(\theta) = -\mathbb{E}_{(x,y_w,y_l) \sim \mathcal D_\textnormal{RM}} \bigl[  \log \sigma(r_\theta(x, y_w) - r_\theta(x, y_l)) \bigr],
\]
where $\mathcal D_\textnormal{RM}$ is the pairwise preference dataset used for RM training. We use out-of-sample pairwise accuracy for hyperparameter tuning (learning rate, gradient accumulation), resulting in a model that reached approximately 57\% out-of-sample pairwise accuracy.

After RM training, we apply the proximal policy optimization (PPO) algorithm \cite{ouyang2022training,schulman2017proximal} to fine-tune the LLM and align it with high-performance ad text. A length penalty was added to the reward to counteract a tendency for a model to generate longer text, leading to the following PPO optimization formulation:
\ifistwocolumn
\begin{align*}
\max_{\pi_\phi} \, \mathbb{E}_{x \sim \mathcal D_\textnormal{LLM}, y\sim \pi_\phi(y|x)} \bigl[ r_\theta(x, y)  &- \beta \, \textnormal{KL} [ \pi_\phi( \cdot \mid x), \pi_\textnormal{ref}( \cdot \mid x)  ] \\
&- \alpha \, \textnormal{length}(y) \bigr],
\end{align*}
\else
\[
\max_{\pi_\phi} \, \mathbb{E}_{x \sim \mathcal D_\textnormal{LLM}, y\sim \pi_\phi(y|x)} \bigl[ r_\theta(x, y)  - \beta \, \textnormal{KL} [ \pi_\phi( \cdot \mid x), \pi_\textnormal{ref}( \cdot \mid x)  ] - \alpha \, \textnormal{length}(y) \bigr],
\]
\fi
where $\pi_\phi$ is the LLM, $\mathcal D_\textnormal{LLM}$ is the pointwise dataset used for LLM post-training, $\textnormal{KL}(p,q)$ is the Kullback-Leibler divergence between two probability distributions $p$ and $q$, $\textnormal{length}(y)$ is the number of tokens in $y$, and $\beta$ and $\alpha$ are weight parameters. We found that PPO can reliably increase the RM score, but subjective text quality started to decrease after a certain number of training steps (as evidenced by irrelevant or repetitive text). This is likely due to \emph{overoptimization}, also known as \emph{reward hacking}, a phenomenon where PPO starts to optimize the imperfections of the RM; see \cite{gao2023scaling}. We deal with overoptimization by carefully selecting a model checkpoint using a combination of two strategies: (1) close monitoring the LLM training process using an \emph{evaluation RM} trained on a different data split and (2) a small-scale human preference labeling.

\subsection{Statistical Analysis}
\subsubsection{Descriptive Statistics and Covariates}
\label{app:table_one}

In Table \ref{table:covariates}, we present descriptive statistics of our sample of advertisers, along with balance of covariates.
\begin{table*}[!ht]
\footnotesize
\centering
\setlength{\tabcolsep}{6pt}

\caption{Descriptive statistics of the advertisers in our A/B test. For categorical variables (is\_business\_account, budget\_cat, expertise\_cat, vertical), group differences were assessed using a $\chi^2$ test. For the remaining variables, two-sample t-tests for equality of means were conducted, and two-sided p-values are reported. We do not observe statistically significant imbalances between the groups, indicating balanced sample characteristics.}

\vspace{10pt}

\begin{tabular}{lrrr}
\textbf{Variable} & \textbf{Control} & \textbf{Treatment} & \textbf{p-value} \\
 & \textit{Imitation LLM v2} & \textit{AdLlama} &  \\

\hline\hline
    $N$ & 17632 & 17217 & \\
        pre\_exp\_ctr (mean (SD)) & 0.027 (0.022) & 0.027 (0.022) & 0.766 \\
        pre\_exp\_engagement (mean (SD)) & 0.329 (2.556) & 0.331 (4.411) & 0.971 \\
        pre\_exp\_impressions (mean (SD)) & 16.497 (144.382) & 15.811 (148.639) & 0.662 \\
        pre\_exp\_ad\_cnt (mean (SD)) & 0.448 (1.945)   & 0.454 (1.878)  &  0.753 \\     
        account\_age\_yr (mean (SD)) & 3.238 (3.483) & 3.276 (3.487) & 0.307 \\
        nov\_feb\_ad\_cnt (mean (SD)) & 2.007 (9.038) & 2.114 (10.362) & 0.304 \\
        nov\_feb\_variant\_cnt (mean (SD)) & 5.983 (28.393) & 6.435 (33.546) & 0.175 \\
        is\_business\_account = 1 (\%) & 13468 (76.38) & 13113 (76.16) & 0.637 \\
        has\_created\_llm\_ad = 1 (\%) & 4604 (26.11) & 4570 (26.54) & 0.366 \\
        is\_new\_advertiser = 1 (\%) & 1778 (10.08) & 1729 (10.04) & 0.912 \\
        budget\_cat (\%) & & & 0.128 \\
        \quad 1.Low & 8313 (47.15) & 8031 (46.65) & \\
        \quad 2.Mid & 1901 (10.78) & 1777 (10.32) & \\
        \quad 3.High & 7418 (42.07) & 7409 (43.03) & \\
        expertise\_cat = 2.High (\%) & 2881 (16.34) & 2710 (15.74) & 0.131 \\
        vertical (\%) & & & 0.117 \\
        \quad Advertising and Marketing & 610 (3.46) & 595 (3.46) & \\
        \quad Automotive & 510 (2.89) & 512 (2.97) & \\
        \quad Business to Business & 353 (2.00) & 437 (2.54) & \\
        \quad Consumer Packaged Goods & 1195 (6.78) & 1130 (6.56) & \\
        \quad Ecommerce & 1681 (9.53) & 1663 (9.66) & \\
        \quad Entertainment and Media & 2330 (13.21) & 2318 (13.46) & \\
        \quad Healthcare, Pharmaceuticals, and Biotech & 778 (4.41) & 716 (4.16) & \\
        \quad Other & 1010 (5.73) & 990 (5.75) & \\
        \quad Professional Services & 2825 (16.02) & 2782 (16.16) & \\
        \quad Publishing & 635 (3.60) & 551 (3.20) & \\
        \quad Restaurants & 454 (2.57) & 409 (2.38) & \\
        \quad Retail & 3218 (18.25) & 3176 (18.45) & \\
        \quad Technology & 438 (2.48) & 417 (2.42) & \\
        \quad Travel & 581 (3.30) & 570 (3.31) & \\
        \quad Unlisted & 1014 (5.75) & 951 (5.52) & \\
\hline\hline
\end{tabular}
\label{table:covariates}
\end{table*}
 
Below, we give details of each of the covariates used in the study.
\begin{itemize}
    \item pre\_exp\_ctr: Lifetime, pre-experiment CTR across all ads launched on Meta apps.
    \item pre\_exp\_engagement: Lifetime, pre-experiment engagement count across all ads launched on Meta apps. Units are in millions.
    \item pre\_exp\_impressions: Lifetime, pre-experiment impression count across all ads launched on Meta apps. Units are in millions.
    \item pre\_exp\_ad\_cnt: Lifetime, pre-experiment ad count launched on all Meta apps. Units are in thousands.
    \item account\_age\_yr: The number of years since the advertiser's account was created.
    \item nov\_feb\_ad\_cnt: The number of ads created by the advertiser during the period after the launch of the initial Text Generation feature, but before the start of the A/B test.
    \item nov\_feb\_variant\_cnt: The number of ad variants created by the advertiser during the period after the launch of the initial Text Generation feature, but before the start of the A/B test.
    \item is\_business\_account: A binary variable indicating whether the advertiser's account is associated with a business page.
    \item has\_created\_llm\_ad: A binary variable indicating whether the advertiser created an LLM-generated ad during the period after the launch of the initial Text Generation feature, but before the start of the A/B test.
    \item is\_new\_advertiser: A binary variable indicating whether the advertiser's ads have combined for fewer than 1{,}000 total impressions on Meta apps (this includes many advertisers who have an account, but have delivered zero ad impressions).
    \item budget\_cat: A categorical variable that groups advertisers based on their budget spend. The possible values are ``1.Low,'' ``2.Mid,'' and `3.High.''
    \item expertise\_cat: A categorical variable that groups advertisers based on their expertise level with Meta's ad system. High expertise means that the advertiser makes use of more advanced ad features. The possible values are ``1.Low'' and ``2.High.''
    \item vertical: A categorical variable indicating the vertical in which the advertiser operates. Examples include ``Retail,'' ``Travel,'' or ``Technology.''
\end{itemize}

\subsubsection{Interpretation of Regression Specification under Log-link Function}
\label{app:loglinkinterp}
Consider the log-binomial regression model specified in (\ref{eq:main_binom}). Exponentiating both sides, we have
\ifistwocolumn
\begin{equation*}
\label{eq:main_binom_interp}
\begin{aligned}
    &\mathbb{E}(Y_i / n_i \, | \, X_i) \\
    &= \begin{aligned}[t]&\exp \bigl ( \beta_2 \cdot \mathbf{1}_{\textnormal{existing}(i)} \cdot \log(\textnormal{CTR}^{\text{pre}}_i\bigr) + \beta_3 \cdot \mathbf{1}_{\textnormal{new}(i)}) \\
    &\cdot \exp(\textnormal{``other covariates''})\end{aligned}\\
    &= \begin{aligned}[t]&\bigl[ \mathbf{1}_{\textnormal{existing}(i)} \, (\textnormal{CTR}^{\text{pre}}_i\bigr)^{\beta_2}  + \mathbf{1}_{\textnormal{new}(i)} \, \exp(\beta_3)\bigr] \\
    &\cdot \exp(\textnormal{``other covariates''})\end{aligned}\\
    &= (\textnormal{``baseline'' CTR}) \, \exp(\textnormal{``other covariates''}).
\end{aligned}
\end{equation*}
\else
\begin{equation*}
\label{eq:main_binom_interp}
\begin{aligned}
    \mathbb{E}(Y_i / n_i \, | \, X_i) &= \exp \bigl ( \beta_2 \cdot \mathbf{1}_{\textnormal{existing}(i)} \cdot \log(\textnormal{CTR}^{\text{pre}}_i\bigr) + \beta_3 \cdot \mathbf{1}_{\textnormal{new}(i)}) \, \exp(\textnormal{``other covariates''})\\
    &= \bigl[ \mathbf{1}_{\textnormal{existing}(i)} \, (\textnormal{CTR}^{\text{pre}}_i\bigr)^{\beta_2}  + \mathbf{1}_{\textnormal{new}(i)} \, \exp(\beta_3)\bigr] \, \exp(\textnormal{``other covariates''})\\
    &= (\textnormal{``baseline'' CTR}) \, \exp(\textnormal{``other covariates''}).
\end{aligned}
\end{equation*}
\fi
The last equality illustrates that we can interpret the term in brackets as an estimated ``baseline'' CTR for advertiser $i$ based on past performance. For existing advertisers (i.e., if $\textnormal{existing}(i)$ is true), then baseline CTR term in brackets reduces to $(\textnormal{CTR}^{\text{pre}}_i\bigr)^{\beta_2}$, where we may interpret $\beta_2$ as an adjustment to convert between lifetime CTR on Meta's apps to CTR on the Facebook platform. On the other hand, for new advertisers (i.e., $\textnormal{new}(i)$ is true), we do not have an observation for their pre-experiment CTR and the term in brackets simply reduces to $\exp(\beta_3)$. This model-based term serves to estimate advertiser $i$'s CTR in the absence of a pre-experiment CTR.

Therefore, we can interpret the entire regression as a ``baseline CTR'' further adjusted by other covariates. This interpretation is possible because we incorporated a log-transformed $\text{CTR}_i^\text{pre}$ into a regression formulation with a log link function.

\subsubsection{Model-free Evidence}
\label{app:model_free}
It is not immediately obvious how to compute the ``average'' CTR across the treatment and control groups---simply computing advertiser-level CTRs and then computing sample averages treats low-impression advertisers and high-impression advertisers equivalently, but intuitively (absent a particular regression model), high-impression advertisers offer more precise CTR observations and should therefore be weighted higher. 

It is common in A/B testing to compute the \emph{global} CTR of each group by computing the ratio of the total number of engagements to the total number of impressions, i.e., $(\sum_i Y_i) / (\sum_i n_i)$. It turns out that global CTR is equivalent to an \emph{impression-weighted} average, which matches our intuition above: $(\sum_i Y_i) / (\sum_i n_i) = \sum_i w_i \, (Y_i / n_i)$, where $w_i = n_i / (\sum_j n_j)$; see \cite{nie2020dealing}. We apply the Delta method to compute the confidence interval around this estimate; see Equation 4 of \cite{deng2018applying} for an example. The model-free estimate is shown in Figure \ref{fig:model_free_ctr}.

\ifistwocolumn
\begin{figure}[!ht]
	\centering
	\includegraphics[scale=0.75]{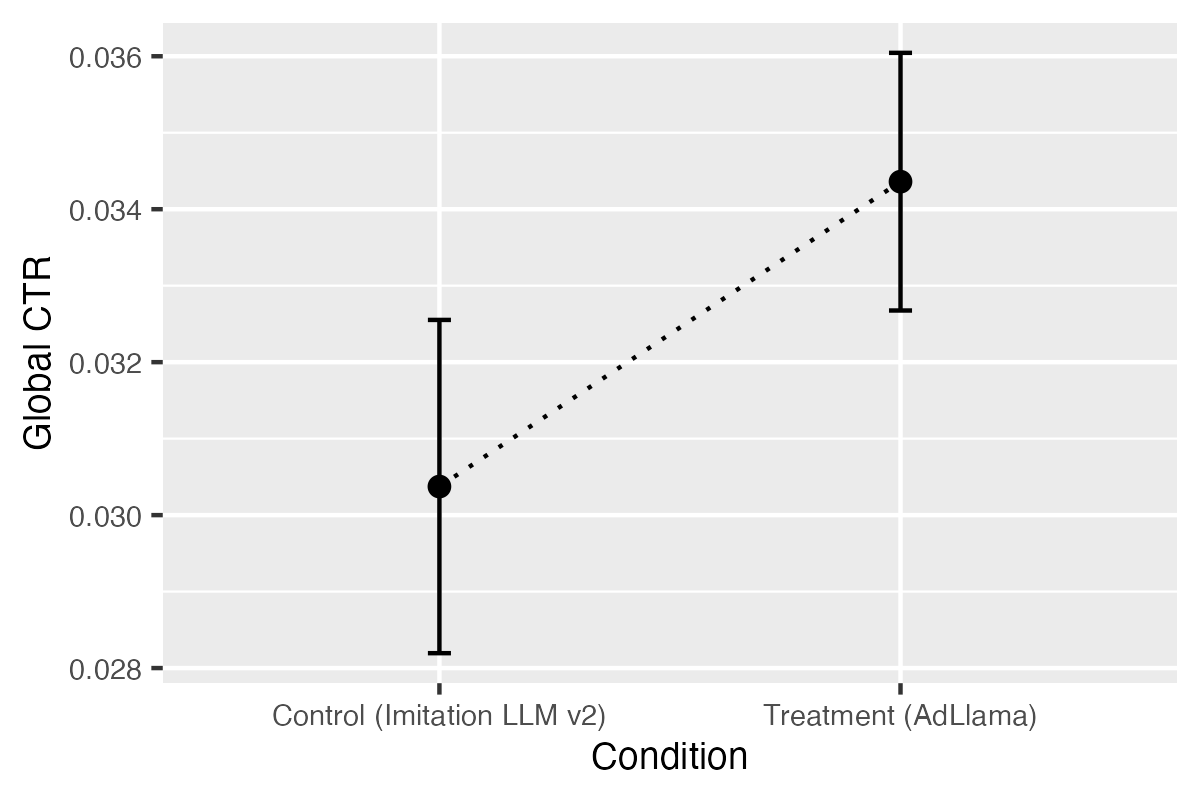} 
	\caption{\textbf{Model-free CTR estimates.} The control group's CTR is estimated at 0.0304, with a 95\% confidence interval of (0.0282, 0.0326). The treatment group's CTR is 0.0344, with a 95\% confidence interval of (0.0327, 0.0360), illustrated by the error bars in the plot. A two-sided z-test indicates a statistically significant difference between the groups ($p=0.0046$).}
\label{fig:model_free_ctr}
\end{figure}
\else
\begin{figure}[!ht]
	\centering
	\includegraphics[scale=1.0]{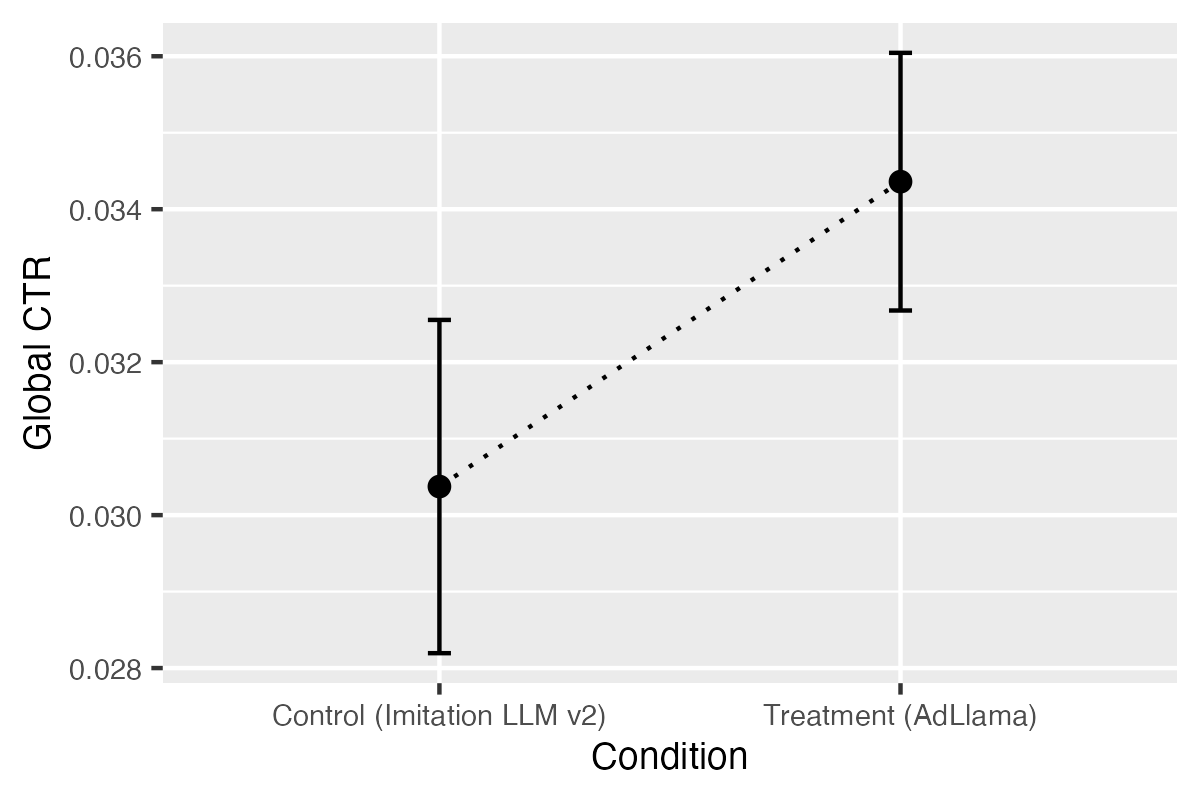} 
	\caption{\textbf{Model-free CTR estimates.} The control group's CTR is estimated at 0.0304, with a 95\% confidence interval of (0.0282, 0.0326). The treatment group's CTR is 0.0344, with a 95\% confidence interval of (0.0327, 0.0360), illustrated by the error bars in the plot. A two-sided z-test indicates a statistically significant difference between the groups ($p=0.0046$).}
\label{fig:model_free_ctr}
\end{figure}
\fi

\subsubsection{Binomial, Logistic, and Poisson Regressions}
\label{app:logistic_pois}
First, the full results of our main log-binomial regression specification are given in Table \ref{table:log_binom_full} (Table \ref{table:log_binom} is an abbreviated version). As an alternative to the log-binomial model of Equation \ref{eq:main_binom}, we also consider a logistic regression specification. The only difference from the log-binomial regression is a logit link function. These results are given in Table \ref{table:logistic}; the results are in agreement with the log-binomial specification.

So far, we have used heteroskedasticity-consistent standard errors in our models to ensure that inference is robust. Another approach is to use quasi-binomial regression, which introduces a dispersion parameter to adjust for variability exceeding that predicted by the binomial distribution, thereby providing more reliable standard error estimates. See Table \ref{table:quasi_binom} for results. We observe that the standard errors under quasi-binomial specification are significantly less conservative than using robust standard errors. Overall, the results are qualitatively similar.

Finally, we consider and Poisson and quasi-Poisson regression specifications, which can be used to estimate \emph{rates} using an ``offset'' term \cite{frome1983analysis}. This is specified as follows:
\ifistwocolumn
\begin{equation}
\label{eq:supp_poisson}
\begin{aligned}
\log \mathbb{E}(Y_i \mid X_i)
&= \beta_0 + \beta_1 \cdot \text{AdLlama}_i \\
&\quad + \beta_2 \cdot \mathbf{1}_{\text{exist}(i)} \cdot \log(\text{CTR}^{\text{pre}}_i) \\
&\quad + \beta_3 \cdot \mathbf{1}_{\text{new}(i)} + \boldsymbol{\beta}_{4{:}10}^T Z_i \\
&\quad + \theta_{\text{vert}(i)} + \alpha_{\text{budget}(i)} + \lambda_{\text{expertise}(i)} \\
&\quad + \log n_i \vphantom{\log(\text{CTR}^{\text{pre}}_i)}.
\end{aligned}
\end{equation}
\else
\begin{equation*}
\label{eq:supp_poisson}
\begin{aligned}
    \log \mathbb{E}(Y_i \, | \, X_i) = \beta_0 &+ \beta_1 \cdot \textnormal{AdLlama}_i + \beta_2 \cdot \mathbf{1}_{\textnormal{existing}(i)} \cdot \log(\textnormal{CTR}^{\text{pre}}_i) + \beta_3 \cdot \mathbf{1}_{\textnormal{new}(i)}  \\
    &  + \boldsymbol{\beta}_{4:10}^T \, \, Z_i + \theta_{\textnormal{vert}(i)} + \alpha_{\textnormal{budget}(i)} + \lambda_{\textnormal{expertise}(i)} + \log n_i.
\end{aligned}
\end{equation*}
\fi
The term $\log n_i$ is called the ``offset'' and its coefficient is constrained to be 1, so that when moved to the left hand side, we obtain $\log ( \mathbb{E}(Y_i \, | \, X_i) / n_i)$, precisely the CTR that we wish to estimate. The results for Poisson regression (with HC1 standard errors) are in Table \ref{table:poisson} and results for the quasi-likelihood variant are in Table \ref{table:quasi_poisson}. The findings are again consistent with the other specifications.

\begin{table*}[!bp] \centering 
    \footnotesize
  \caption{The full version of Table \ref{table:log_binom} for the log-binomial regression; includes fixed effects.} 
  \label{table:log_binom_full} 
\begin{tabular}{@{\extracolsep{5pt}}lc} 
\\[-1.8ex]\hline 
\hline \\[-1.8ex] 
 & \multicolumn{1}{c}{\textit{Dependent variable:}} \\ 
\cline{2-2} 
\\[-1.8ex] & engagement/impressions \\ 
\hline \\[-1.8ex] 
 \textbf{treatment} & \textbf{0.0651}$^{**}$ \textbf{(0.0299)} \\ 
  log\_pre\_exp\_ctr\_existing & 0.5931$^{***}$ (0.0264) \\ 
  is\_new\_advertiser & $-$2.3469$^{***}$ (0.1569) \\ 
  pre\_exp\_ad\_cnt & 0.0024 (0.0028) \\ 
  pre\_exp\_impressions & $-$0.0001 (0.0001) \\ 
  pre\_exp\_engagement & 0.0036 (0.0049) \\ 
  account\_age\_yr & 0.0008 (0.0038) \\ 
  nov\_feb\_ad\_cnt & 0.0013 (0.0019) \\ 
  nov\_feb\_variant\_cnt & $-$0.0003 (0.0006) \\ 
  is\_business\_account & $-$0.0286 (0.0424) \\ 
  has\_created\_llm\_ad & $-$0.0109 (0.0345) \\ 
  budget\_cat: 2.Mid & 0.0460 (0.0508) \\ 
  budget\_cat: 3.High & $-$0.0237 (0.0417) \\ 
  expertise\_cat: 2.High & 0.0005 (0.0338) \\ 
  vertical: Automotive & $-$0.1556 (0.1219) \\ 
  vertical: Business to Business & $-$0.4097$^{**}$ (0.1867) \\ 
  vertical: Consumer Packaged Goods & $-$0.1476 (0.1161) \\ 
  vertical: Ecommerce & $-$0.0847 (0.1149) \\ 
  vertical: Entertainment and Media & 0.0648 (0.1134) \\ 
  vertical: Healthcare, Pharmaceuticals, and Biotech & $-$0.1418 (0.1233) \\ 
  vertical: Other & $-$0.2076 (0.1275) \\ 
  vertical: Professional Services & $-$0.1612 (0.1248) \\ 
  vertical: Publishing & $-$0.0107 (0.1451) \\ 
  vertical: Restaurants & 0.0117 (0.1185) \\ 
  vertical: Retail & $-$0.1914 (0.1179) \\ 
  vertical: Technology & $-$0.0369 (0.1360) \\ 
  vertical: Travel & 0.0163 (0.1179) \\ 
  vertical: Unlisted & 0.1067 (0.1681) \\ 
  Constant & $-$1.1233$^{***}$ (0.1506) \\ 
 \hline \\[-1.8ex] 
Observations & 34,849 \\ 
\hline 
\hline \\[-1.8ex] 
\textit{Note:} HC1 robust standard errors in parentheses.  & \multicolumn{1}{r}{$^{*}$p$<$0.1; $^{**}$p$<$0.05; $^{***}$p$<$0.01} \\ 
\end{tabular} 
\end{table*}

\begin{table*}[!bp] \centering 
    \footnotesize
  \caption{Logistic regression specification results. We report HC1 robust standard errors and two-sided p-values.} 
  \label{table:logistic} 
\begin{tabular}{@{\extracolsep{5pt}}lc} 
\\[-1.8ex]\hline 
\hline \\[-1.8ex] 
 & \multicolumn{1}{c}{\textit{Dependent variable:}} \\ 
\cline{2-2} 
\\[-1.8ex] & engagement/impressions \\ 
\hline \\[-1.8ex] 
 \textbf{treatment} & \textbf{0.0685$^{**}$ (0.0311)} \\ 
  log\_pre\_exp\_ctr\_existing & 0.6158$^{***}$ (0.0277) \\ 
  is\_new\_advertiser & $-$2.4418$^{***}$ (0.1631) \\ 
  pre\_exp\_ad\_cnt & 0.0024 (0.0029) \\ 
  pre\_exp\_impressions & $-$0.0001 (0.0001) \\ 
  pre\_exp\_engagement & 0.0036 (0.0052) \\ 
  account\_age\_yr & 0.0009 (0.0040) \\ 
  nov\_feb\_ad\_cnt & 0.0013 (0.0020) \\ 
  nov\_feb\_variant\_cnt & $-$0.0003 (0.0006) \\ 
  is\_business\_account & $-$0.0318 (0.0444) \\ 
  has\_created\_llm\_ad & $-$0.0118 (0.0357) \\ 
  budget\_cat: 2.Mid & 0.0484 (0.0528) \\ 
  budget\_cat: 3.High & $-$0.0271 (0.0433) \\ 
  expertise\_cat: 2.High & 0.0006 (0.0351) \\ 
  vertical: Automotive & $-$0.1572 (0.1275) \\ 
  vertical: Business to Business & $-$0.4172$^{**}$ (0.1914) \\ 
  vertical: Consumer Packaged Goods & $-$0.1497 (0.1213) \\ 
  vertical: Ecommerce & $-$0.0848 (0.1202) \\ 
  vertical: Entertainment and Media & 0.0736 (0.1187) \\ 
  vertical: Healthcare, Pharmaceuticals, and Biotech & $-$0.1415 (0.1286) \\ 
  vertical: Other & $-$0.2132 (0.1330) \\ 
  vertical: Professional Services & $-$0.1641 (0.1302) \\ 
  vertical: Publishing & $-$0.0044 (0.1518) \\ 
  vertical: Restaurants & 0.0178 (0.1239) \\ 
  vertical: Retail & $-$0.1952 (0.1232) \\ 
  vertical: Technology & $-$0.0347 (0.1420) \\ 
  vertical: Travel & 0.0207 (0.1233) \\ 
  vertical: Unlisted & 0.1156 (0.1747) \\ 
  Constant & $-$1.0005$^{***}$ (0.1584) \\ 
 \hline \\[-1.8ex] 
Observations & 34,849 \\ 
\hline 
\hline \\[-1.8ex] 
\textit{Note:} HC1 robust standard errors in parentheses.  & \multicolumn{1}{r}{$^{*}$p$<$0.1; $^{**}$p$<$0.05; $^{***}$p$<$0.01} \\ 
\end{tabular} 
\end{table*}

\begin{table*}[!p] \centering 
  \footnotesize
  \caption{Quasi-binomial regression specification results for both log and logit link functions. The fitted dispersion parameters are 8.427 and 8.433, respectively. We report two-sided p-values.} 
  \label{table:quasi_binom} 
\begin{tabular}{@{\extracolsep{5pt}}lcc} 
\\[-1.8ex]\hline 
\hline \\[-1.8ex] 
 & \multicolumn{2}{c}{\textit{Dependent variable:}} \\ 
\cline{2-3} 
\\[-1.8ex] & \multicolumn{2}{c}{engagement/impressions} \\ 
 & \textit{link = log} & \textit{link = logit} \\ 
\hline \\[-1.8ex] 
 \textbf{treatment} & \textbf{0.0651}$^{***}$ \textbf{(0.0073)} & \textbf{0.0685}$^{***}$ \textbf{(0.0076)} \\ 
  log\_pre\_exp\_ctr\_existing & 0.5931$^{***}$ (0.0069) & 0.6158$^{***}$ (0.0073) \\ 
  is\_new\_advertiser & $-$2.3469$^{***}$ (0.0339) & $-$2.4418$^{***}$ (0.0354) \\ 
  pre\_exp\_ad\_cnt & 0.0024$^{*}$ (0.0012) & 0.0024$^{*}$ (0.0012) \\ 
  pre\_exp\_impressions & $-$0.0001$^{***}$ (0.00003) & $-$0.0001$^{***}$ (0.00003) \\ 
  pre\_exp\_engagement & 0.0036$^{***}$ (0.0011) & 0.0036$^{***}$ (0.0011) \\ 
  account\_age\_yr & 0.0008 (0.0010) & 0.0009 (0.0011) \\ 
  nov\_feb\_ad\_cnt & 0.0013$^{*}$ (0.0007) & 0.0013$^{*}$ (0.0007) \\ 
  nov\_feb\_variant\_cnt & $-$0.0003 (0.0002) & $-$0.0003 (0.0002) \\ 
  is\_business\_account & $-$0.0286$^{**}$ (0.0115) & $-$0.0318$^{***}$ (0.0120) \\ 
  has\_created\_llm\_ad & $-$0.0109 (0.0085) & $-$0.0118 (0.0088) \\ 
  budget\_cat2.Mid & 0.0460$^{***}$ (0.0165) & 0.0484$^{***}$ (0.0172) \\ 
  budget\_cat3.High & $-$0.0237$^{**}$ (0.0105) & $-$0.0271$^{**}$ (0.0109) \\ 
  expertise\_cat2.High & 0.0005 (0.0091) & 0.0006 (0.0095) \\ 
  vertical: Automotive & $-$0.1556$^{***}$ (0.0298) & $-$0.1572$^{***}$ (0.0310) \\ 
  vertical: Business to Business & $-$0.4097$^{***}$ (0.0316) & $-$0.4172$^{***}$ (0.0327) \\ 
  vertical: Consumer Packaged Goods & $-$0.1476$^{***}$ (0.0257) & $-$0.1497$^{***}$ (0.0268) \\ 
  vertical: Ecommerce & $-$0.0847$^{***}$ (0.0250) & $-$0.0848$^{***}$ (0.0261) \\ 
  vertical: Entertainment and Media & 0.0648$^{***}$ (0.0244) & 0.0736$^{***}$ (0.0255) \\ 
  vertical: Healthcare, Pharmaceuticals, and Biotech & $-$0.1418$^{***}$ (0.0323) & $-$0.1415$^{***}$ (0.0337) \\ 
  vertical: Other & $-$0.2076$^{***}$ (0.0291) & $-$0.2132$^{***}$ (0.0302) \\ 
  vertical: Professional Services & $-$0.1612$^{***}$ (0.0260) & $-$0.1641$^{***}$ (0.0270) \\ 
  vertical: Publishing & $-$0.0107 (0.0274) & $-$0.0044 (0.0286) \\ 
  vertical: Restaurants & 0.0117 (0.0333) & 0.0178 (0.0349) \\ 
  vertical: Retail & $-$0.1914$^{***}$ (0.0248) & $-$0.1952$^{***}$ (0.0258) \\ 
  vertical: Technology & $-$0.0369 (0.0308) & $-$0.0347 (0.0319) \\ 
  vertical: Travel & 0.0163 (0.0278) & 0.0207 (0.0290) \\ 
  vertical: Unlisted & 0.1067$^{***}$ (0.0390) & 0.1156$^{***}$ (0.0405) \\ 
  Constant & $-$1.1233$^{***}$ (0.0355) & $-$1.0005$^{***}$ (0.0373) \\ 
 \hline \\[-1.8ex] 
Observations & 34,849 & 34,849 \\ 
\hline 
\hline \\[-1.8ex] 
\textit{Note:} Standard errors adjusted by the dispersion parameter.  & \multicolumn{2}{r}{$^{*}$p$<$0.1; $^{**}$p$<$0.05; $^{***}$p$<$0.01} \\ 
\end{tabular} 
\end{table*}

\begin{table*}[!p] \centering 
  \footnotesize
  \caption{Poisson regression results using log impressions as an offset; see Equation \ref{eq:supp_poisson}. We report HC1 robust standard errors and two-sided p-values.} 
  \label{table:poisson} 
\begin{tabular}{@{\extracolsep{5pt}}lc} 
\\[-1.8ex]\hline 
\hline \\[-1.8ex] 
 & \multicolumn{1}{c}{\textit{Dependent variable:}} \\ 
\cline{2-2} 
\\[-1.8ex] & engagement/impressions \\ 
\hline \\[-1.8ex] 
 \textbf{treatment} & \textbf{0.0658}$^{**}$ \textbf{(0.0299)} \\ 
  log\_pre\_exp\_ctr\_existing & 0.5913$^{***}$ (0.0263) \\ 
  is\_new\_advertiser & $-$2.3431$^{***}$ (0.1569) \\ 
  pre\_exp\_ad\_cnt & 0.0023 (0.0028) \\ 
  pre\_exp\_impressions & $-$0.0001 (0.0001) \\ 
  pre\_exp\_engagement & 0.0036 (0.0049) \\ 
  account\_age\_yr & 0.0008 (0.0038) \\ 
  nov\_feb\_ad\_cnt & 0.0013 (0.0019) \\ 
  nov\_feb\_variant\_cnt & $-$0.0003 (0.0006) \\ 
  is\_business\_account & $-$0.0299 (0.0425) \\ 
  has\_created\_llm\_ad & $-$0.0110 (0.0344) \\ 
  budget\_cat: 2.Mid & 0.0461 (0.0506) \\ 
  budget\_cat: 3.High & $-$0.0259 (0.0417) \\ 
  expertise\_cat2.High & 0.0001 (0.0338) \\ 
  vertical: Automotive & $-$0.1515 (0.1224) \\ 
  vertical: Business to Business & $-$0.4075$^{**}$ (0.1862) \\ 
  vertical: Consumer Packaged Goods & $-$0.1442 (0.1165) \\ 
  vertical: Ecommerce & $-$0.0812 (0.1153) \\ 
  vertical: Entertainment and Media & 0.0705 (0.1137) \\ 
  vertical: Healthcare, Pharmaceuticals, and Biotech & $-$0.1364 (0.1233) \\ 
  vertical: Other & $-$0.2064 (0.1277) \\ 
  vertical: Professional Services & $-$0.1581 (0.1253) \\ 
  vertical: Publishing & $-$0.0045 (0.1451) \\ 
  vertical: Restaurants & 0.0172 (0.1187) \\ 
  vertical: Retail & $-$0.1868 (0.1182) \\ 
  vertical: Technology & $-$0.0330 (0.1366) \\ 
  vertical: Travel & 0.0206 (0.1182) \\ 
  vertical: Unlisted & 0.1119 (0.1686) \\ 
  Constant & $-$1.1313$^{***}$ (0.1508) \\ 
 \hline \\[-1.8ex] 
Observations & 34,849 \\ 
\hline 
\hline \\[-1.8ex] 
\textit{Note:} HC1 robust standard errors in parentheses.  & \multicolumn{1}{r}{$^{*}$p$<$0.1; $^{**}$p$<$0.05; $^{***}$p$<$0.01} \\ 
\end{tabular} 
\end{table*}

\begin{table*}[!htbp] \centering \footnotesize
  \caption{Quasi-Poisson regression specification results. The fitted dispersion parameter is 8.127. We report two-sided p-values.} 
  \label{table:quasi_poisson} 
\begin{tabular}{@{\extracolsep{5pt}}lc} 
\\[-1.8ex]\hline 
\hline \\[-1.8ex] 
 & \multicolumn{1}{c}{\textit{Dependent variable:}} \\ 
\cline{2-2} 
\\[-1.8ex] & engagement/impressions \\ 
\hline \\[-1.8ex] 
 \textbf{treatment} & \textbf{0.0658}$^{***}$ \textbf{(0.0073)} \\ 
  log\_pre\_exp\_ctr\_existing & 0.5913$^{***}$ (0.0070) \\ 
  is\_new\_advertiser & $-$2.3431$^{***}$ (0.0340) \\ 
  pre\_exp\_ad\_cnt & 0.0023$^{*}$ (0.0012) \\ 
  pre\_exp\_impressions & $-$0.0001$^{***}$ (0.00003) \\ 
  pre\_exp\_engagement & 0.0036$^{***}$ (0.0011) \\ 
  account\_age\_yr & 0.0008 (0.0010) \\ 
  nov\_feb\_ad\_cnt & 0.0013$^{*}$ (0.0007) \\ 
  nov\_feb\_variant\_cnt & $-$0.0003 (0.0002) \\ 
  is\_business\_account & $-$0.0299$^{***}$ (0.0115) \\ 
  has\_created\_llm\_ad & $-$0.0110 (0.0085) \\ 
  budget\_cat2.Mid & 0.0461$^{***}$ (0.0166) \\ 
  budget\_cat3.High & $-$0.0259$^{**}$ (0.0105) \\ 
  expertise\_cat2.High & 0.0001 (0.0091) \\ 
  vertical: Automotive & $-$0.1515$^{***}$ (0.0299) \\ 
  vertical: Business to Business & $-$0.4075$^{***}$ (0.0316) \\ 
  vertical: Consumer Packaged Goods & $-$0.1442$^{***}$ (0.0258) \\ 
  vertical: Ecommerce & $-$0.0812$^{***}$ (0.0251) \\ 
  vertical: Entertainment and Media & 0.0705$^{***}$ (0.0245) \\ 
  vertical: Healthcare, Pharmaceuticals, and Biotech & $-$0.1364$^{***}$ (0.0324) \\ 
  vertical: Other & $-$0.2064$^{***}$ (0.0291) \\ 
  vertical: Professional Services & $-$0.1581$^{***}$ (0.0261) \\ 
  vertical: Publishing & $-$0.0045 (0.0275) \\ 
  vertical: Restaurants & 0.0172 (0.0335) \\ 
  vertical: Retail & $-$0.1868$^{***}$ (0.0249) \\ 
  vertical: Technology & $-$0.0330 (0.0308) \\ 
  vertical: Travel & 0.0206 (0.0279) \\ 
  vertical: Unlisted & 0.1119$^{***}$ (0.0391) \\ 
  Constant & $-$1.1313$^{***}$ (0.0357) \\ 
 \hline \\[-1.8ex] 
Observations & 34,849 \\ 
\hline 
\hline \\[-1.8ex] 
\textit{Note:} Standard errors adjusted by the dispersion parameter.  & \multicolumn{1}{r}{$^{*}$p$<$0.1; $^{**}$p$<$0.05; $^{***}$p$<$0.01} \\ 
\end{tabular} 
\end{table*}

\subsubsection{Separate Engagement and Impression Linear Regressions}
\label{app:separate}
Alternatively, we can consider a regression specification where engagement and impressions are modeled separately as linear regressions, using the same covariates as we did above:
\ifistwocolumn
\begin{equation}
\label{eq:supp_outcome}
\begin{aligned}
\text{Outcome}_i 
&= \beta_0 + \beta_1 \cdot \text{AdLlama}_i \\
&\quad + \beta_2 \cdot \mathbf{1}_{\text{exist}(i)} \cdot \text{CTR}^{\text{pre}}_i \\
&\quad + \beta_3 \cdot \mathbf{1}_{\text{new}(i)} + \boldsymbol{\beta}_{4{:}10}^T Z_i \\
&\quad + \theta_{\text{vert}(i)} + \alpha_{\text{budget}(i)} + \lambda_{\text{expertise}(i)} + \epsilon_i \vphantom{\text{CTR}^{\text{pre}}_i}.
\end{aligned}
\end{equation}
\else
\begin{equation*}
\begin{aligned}
    \text{Outcome}_i = \beta_0 &+ \beta_1 \cdot \textnormal{AdLlama}_i + \beta_2 \cdot \mathbf{1}_{\textnormal{existing}(i)} \cdot \textnormal{CTR}^{\text{pre}}_i + \beta_3 \cdot \mathbf{1}_{\textnormal{new}(i)}  \\
    &  + \boldsymbol{\beta}_{4:10}^T \, \, Z_i + \theta_{\textnormal{vert}(i)} + \alpha_{\textnormal{budget}(i)} + \lambda_{\textnormal{expertise}(i)} + \epsilon_i,
\end{aligned}
\end{equation*}
\fi
where $\text{Outcome}_i$ refers to either advertiser $i$'s engagement count or impression count and $\epsilon_i$ is the error term. The results are given in Table \ref{table:lin_eng_imp}, where we find a statistically significant increase in engagement (+4.068, $p=0.0023$) when using AdLlama, but no evidence for a change in impressions. Increased engagement while impressions are held constant is consistent with the result of increased CTR from the log-binomial regression of Table \ref{table:log_binom}.
\begin{table*}[!htbp] \centering 
  \footnotesize
  \caption{Separate linear regression results on engagements and impressions. All other notation remains consistent with Table \ref{table:log_binom}. We report HC1 robust standard errors.} 
  \label{table:lin_eng_imp} 
\begin{tabular}{@{\extracolsep{5pt}}lcc} 
\\[-1.8ex]\hline 
\hline \\[-1.8ex] 
 & \multicolumn{2}{c}{\textit{Dependent variable:}} \\ 
\cline{2-3} 
\\[-1.8ex] & engagement & impressions \\ 
\\[-4.8ex] & & \\ 
\hline \\[-1.8ex] 
 \textbf{treatment} & \textbf{4.068}$^{***}$ \textbf{(1.334)} & \textbf{54.637 (49.730)} \\ 
  pre\_exp\_ctr\_existing & 175.495$^{***}$ (31.954) & $-$1,982.590$^{***}$ (692.366) \\ 
  pre\_exp\_ad\_cnt & 1.248 (0.937) & 44.755 (31.544) \\ 
  pre\_exp\_engagement & 0.407 (1.151) & $-$37.671 (31.945) \\ 
  pre\_exp\_impressions & 0.032$^{*}$ (0.019) & 2.686$^{***}$ (0.794) \\ 
  account\_age\_yr & 0.671$^{***}$ (0.227) & 21.168$^{***}$ (8.094) \\ 
  nov\_feb\_ad\_cnt & 0.255 (0.470) & $-$0.976 (13.499) \\ 
  nov\_feb\_variant\_cnt & 0.081 (0.156) & 4.943 (4.559) \\ 
  is\_new\_advertiser & 8.978$^{***}$ (2.843) & 174.461 (207.469) \\ 
  is\_business\_account & 5.757$^{***}$ (1.143) & 183.652$^{***}$ (47.473) \\ 
  has\_created\_llm\_ad & 5.677$^{***}$ (1.847) & 191.082$^{***}$ (61.552) \\ 
  budget\_cat: 2.Mid & 0.553 (1.557) & $-$32.907 (40.244) \\ 
  budget\_cat: 3.High & 14.228$^{***}$ (1.603) & 396.337$^{***}$ (54.707) \\ 
  expertise\_cat: 2.High & $-$1.183 (2.306) & 21.213 (78.768) \\ 
  vertical: Automotive & 8.504$^{**}$ (4.031) & 349.443$^{**}$ (162.753) \\ 
  vertical: Business to Business & 9.267 (7.137) & 847.067 (548.643) \\ 
  vertical: Consumer Packaged Goods & 4.400 (4.069) & 249.672 (153.272) \\ 
  vertical: Ecommerce & 7.378$^{*}$ (3.823) & 265.690$^{*}$ (138.991) \\ 
  vertical: Entertainment and Media & 12.558$^{***}$ (3.645) & 295.972$^{**}$ (129.309) \\ 
  vertical: Healthcare, Pharmaceuticals, and Biotech & 1.378 (3.413) & 57.893 (130.343) \\ 
  vertical: Other & 2.752 (3.158) & 170.415 (124.527) \\ 
  vertical: Professional Services & $-$0.467 (3.041) & 43.760 (124.657) \\ 
  vertical: Publishing & 17.857$^{***}$ (6.602) & 410.544$^{**}$ (208.360) \\ 
  vertical: Restaurants & 3.119 (3.548) & 75.335 (125.864) \\ 
  vertical: Retail & 4.415 (3.174) & 197.073 (126.678) \\ 
  vertical: Technology & 9.659 (8.157) & 376.426 (309.201) \\ 
  vertical: Travel & 14.820$^{***}$ (4.668) & 406.783$^{**}$ (166.148) \\ 
  vertical: Unlisted & 3.004 (3.623) & 17.970 (190.378) \\ 
  Constant & $-$11.150$^{***}$ (3.526) & $-$140.400 (140.123) \\ 
 \hline \\[-1.8ex] 
Observations & 34,849 & 34,849 \\ 
R$^{2}$ & 0.017 & 0.015 \\ 
Adjusted R$^{2}$ & 0.016 & 0.015 \\ 
\hline 
\hline \\[-1.8ex] 
\textit{Note:} HC1 robust standard errors in parentheses.  & \multicolumn{2}{r}{$^{*}$p$<$0.1; $^{**}$p$<$0.05; $^{***}$p$<$0.01} \\ 
\end{tabular} 
\end{table*}

\subsection{Imitation LLM Training Details}
\label{app:imitation_llm}
Imitation LLM v1 and v2 were developed internally at Meta prior to the RLPF work described in this paper. Here, we give a brief description of the training process for these models.
\begin{itemize}
    \item Both of these models used supervised fine-tuning (SFT), with the goal of imitating ``good'' ad text. This follows the instruction-tuning paradigm \cite{ouyang2022training}, where each row of training data is of the form (input ad text, target ad text). The input ad text represents the advertiser's original ad text and the target ad text is a rewritten variation (i.e., an ``improved'' version). Running SFT on this data makes shifts the LLM's response distribution to be more likely to generate the target ad text; in other words, ``imitating'' it.
    \item For Imitation LLM v1, based on 7B \emph{Llama 2 Chat}, the target ad texts were synthetically distilled from the 70B \emph{Llama 2 Chat} model, a more capable but larger model \cite{touvron2023llama2}. Instructions used to prompt the larger model are common ones used in ad copywriting, such as 
    ``paraphrase and shorten'', ``make clear'', ``make actionable'', ``empathize'', ``pose as a question'', or ``focus selling point.''
    \item For Imitation LLM v2, in addition to synthetically distilled target ad texts, human-written ad texts were also added into the mix. This data was collected using similar instructions as for Imitation LLM v1. The synthetic dataset typically offers more creativity, but suffers from higher hallucination rates. On the other hand, the human-rewritten dataset oftentimes has higher quality, but may lack diversity. By training on both, Imitation LLM v2 struck a balance between  creativity and quality.
\end{itemize}

\end{document}